\documentclass{article}

\usepackage{microtype}
\usepackage{graphicx}
\usepackage{subfigure}
\usepackage{booktabs} % for professional tables

\usepackage{hyperref}

\usepackage[accepted]{icml2024}

\usepackage{hyperref} 
\usepackage{url}
\usepackage{epsfig}
\usepackage{graphicx} 
\usepackage{amsmath}
\usepackage{amssymb}
\usepackage{adjustbox}
\usepackage{multirow}
\usepackage{makecell}
\usepackage{multicol}
\usepackage{mathrsfs}
\usepackage{cellspace}
\usepackage{makecell}
\usepackage{bm} 
\usepackage{paralist}
\usepackage{booktabs}
\usepackage{xcolor}
\usepackage{bbding}

\newcommand{\smodel}{CogVLM}
\definecolor{trainablecolor}{RGB}{10,54,164}
\usepackage{wrapfig}

\usepackage{amsmath}
\usepackage{amssymb}
\usepackage{mathtools}
\usepackage{amsthm}

\usepackage[capitalize,noabbrev]{cleveref}

\theoremstyle{plain}

\theoremstyle{definition}

\theoremstyle{remark}

\usepackage[textsize=tiny]{todonotes}

\icmltitlerunning{CogVLM: Visual Expert for Pretrained Language Models}

\begin{document}

\twocolumn[
\icmltitle{CogVLM: Visual Expert for Pretrained Language Models}



\icmlsetsymbol{equal}{*}

\begin{icmlauthorlist}
\icmlauthor{Weihan Wang}{equal,thu}\hspace{-3pt}$^\ddagger$\hspace{3pt}
\icmlauthor{Qingsong Lv}{equal,zp}
\icmlauthor{Wenmeng Yu}{zp}
\icmlauthor{Wenyi Hong}{thu}\hspace{-3pt}$^\ddagger$\hspace{3pt}
\icmlauthor{Ji Qi}{thu}\hspace{-3pt}$^\ddagger$\hspace{3pt}
\icmlauthor{Yan Wang}{zp}
\icmlauthor{Junhui Ji}{zp}
\icmlauthor{Zhuoyi Yang}{thu}\hspace{-3pt}$^\ddagger$\hspace{3pt}
\icmlauthor{Lei Zhao}{zp}
\icmlauthor{Xixuan Song}{thu}\hspace{-3pt}$^\ddagger$\hspace{3pt}
\icmlauthor{Jiazheng Xu}{thu}\hspace{-3pt}$^\ddagger$\hspace{3pt}
\icmlauthor{Keqin Chen}{bh}\hspace{-3pt}$^\ddagger$\hspace{3pt}
\icmlauthor{Bin Xu}{thu}
\icmlauthor{Juanzi Li}{thu}
\icmlauthor{Yuxiao Dong}{thu}
\icmlauthor{Ming Ding}{zp}
\icmlauthor{Jie Tang}{thu}
% \\ \vspace{5pt}
% \icmlauthor{\normalfont wangwh21@mails.tsinghua.edu.cn,}{}
% \icmlauthor{\normalfont ming.ding@zhipuai.cn}{}
\end{icmlauthorlist}

\icmlaffiliation{zp}{Zhipu AI $^\ddagger$Done as intern at Zhipu AI}
\icmlaffiliation{thu}{Tsinghua University}
\icmlaffiliation{bh}{Beihang University}

\icmlcorrespondingauthor{Ming Ding}{ming.ding@zhipuai.cn}
\icmlcorrespondingauthor{Jie Tang}{ jietang@tsinghua.edu.cn}

\icmlkeywords{Multimodal Learning, Representation Learning, Vision and Language}

\vskip 0.3in
]



\printAffiliationsAndNotice{\icmlEqualContribution} % otherwise use the standard text.

\begin{abstract}
We introduce CogVLM, a powerful open-source visual language foundation model.
% a simple and powerful framework to train a visual language model from a pretrained language model. 
Different from the popular \emph{shallow alignment} method which maps image features into the input space of language model, CogVLM bridges the gap between the frozen pretrained language model and image encoder by a trainable visual expert module in the attention and FFN layers. As a result, CogVLM enables a deep fusion of vision language features without sacrificing any performance on NLP tasks. 
% Our model Visual-LLaMA2, which is trained from LLaMA2-7B, achieves state-of-the-art performance on \dm{n} classic cross-modal benchmarks, including \dm{}, surpassing PALI-X 55B. 
% We trained VisualGLM2-en and VisualGLM2-zh from LLaMA2-7B and GLM2-12B respectively, and our models
CogVLM-17B achieves state-of-the-art performance on 17 classic cross-modal benchmarks, including 1) image captioning datasets: NoCaps, Flicker30k, 2) VQA datasets: OKVQA, TextVQA, OCRVQA, ScienceQA, 3) LVLM benchmarks: MM-Vet, MMBench, SEED-Bench, LLaVABench, POPE, MMMU, MathVista, 4) visual grounding datasets: RefCOCO, RefCOCO+, RefCOCOg, Visual7W. Codes and checkpoints are available at https://github.com/THUDM/CogVLM.
% including NoCaps, Flicker30k captioning, OK-VQA, TextVQA, OCRVQA, ScienceQA, TDIUC, MM-Vet, SEED-Bench, MMBench, LLaVA-Bench, POPE, MMMU, MathVista, RefCOCO, RefCOCO+, RefCOCOg and Visual7W. 
\end{abstract}

\begin{figure}[h]
    \centering
    \includegraphics[width=\linewidth]{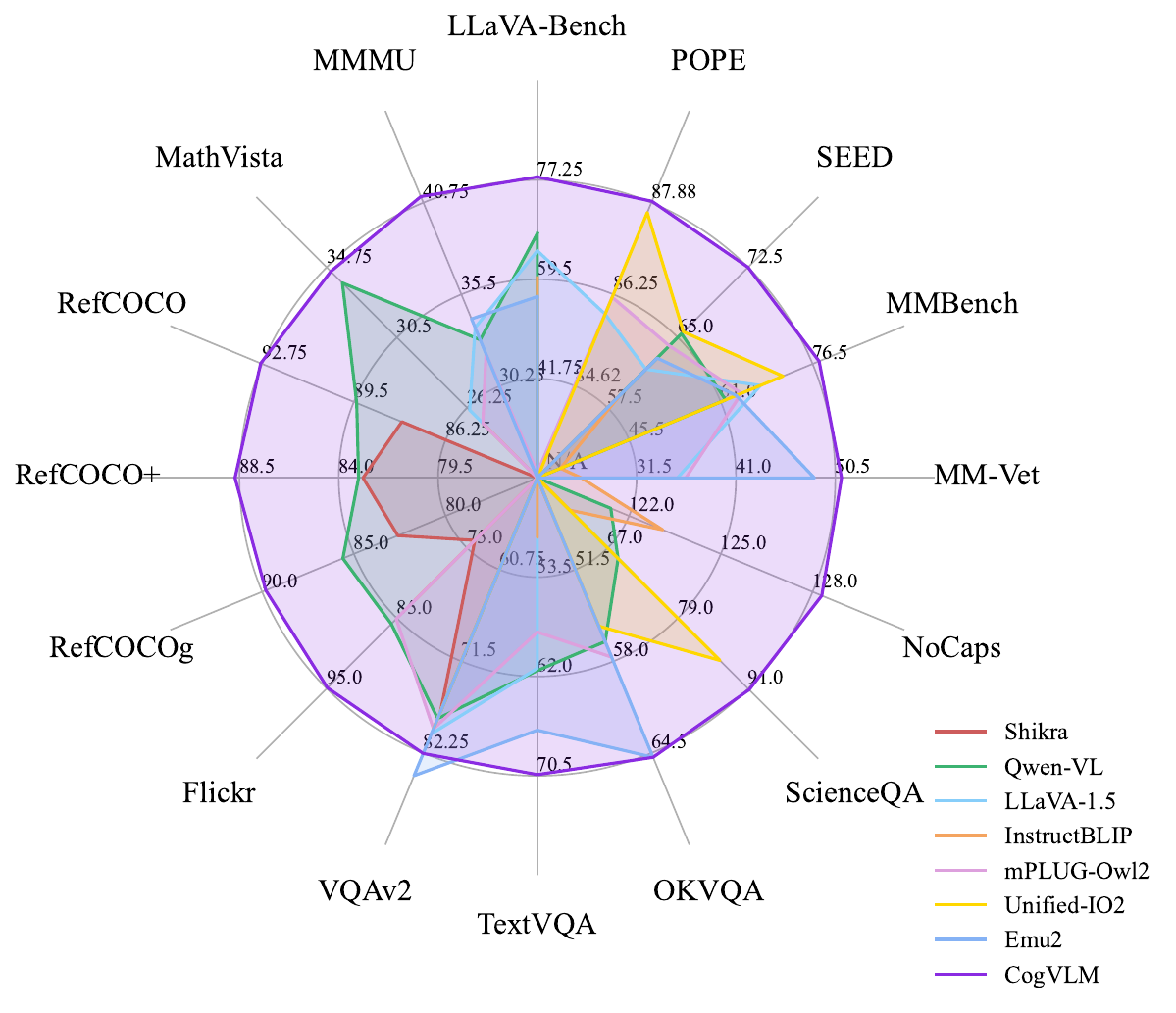}
    \caption{The performance of CogVLM on a broad range of multi-modal tasks in comparison with existing approaches.}
    \label{fig:enter-label}
\end{figure}
% change code: https://github.com/qijimrc/mm_evaluation/blob/master/mmbench/utils/visualization.ipynb

\section{Introduction}
\label{sec:intro}
% Visual language models (VLMs) offer a universal solution for various cross-modal tasks, e.g., image captioning, visual question answering and visual grounding.

\begin{figure*}[]
    \vspace*{-0.5cm}
    \centering
    \includegraphics[width=\linewidth]{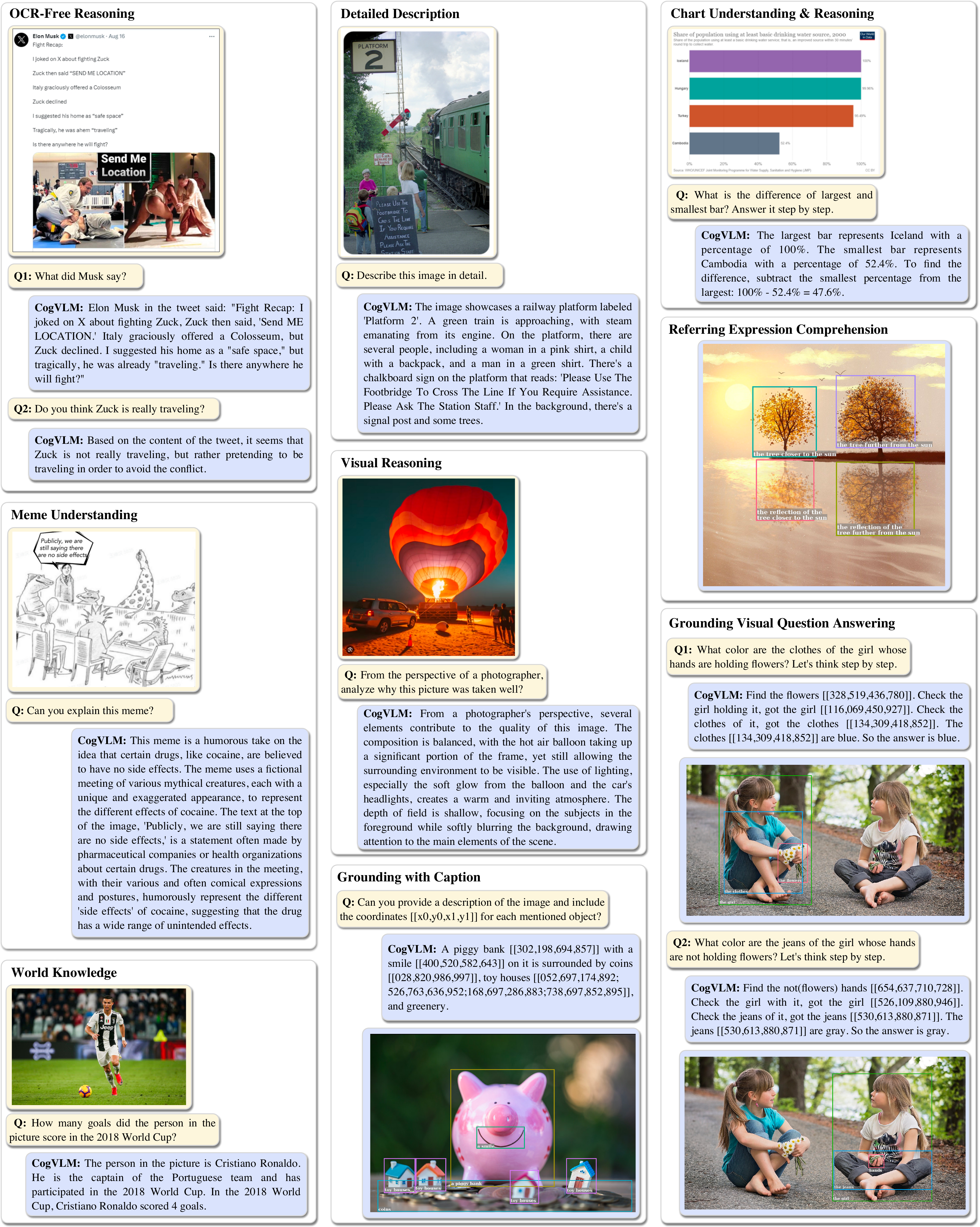}
    \vspace*{-0.3cm}
    \caption{Samples generated by CogVLM.}
    \label{fig:cht}
    \vspace*{-0.6cm}
\end{figure*}

\begin{figure*}[h]
  \centering
  \begin{minipage}{0.48\textwidth}
    \includegraphics[width=\linewidth]{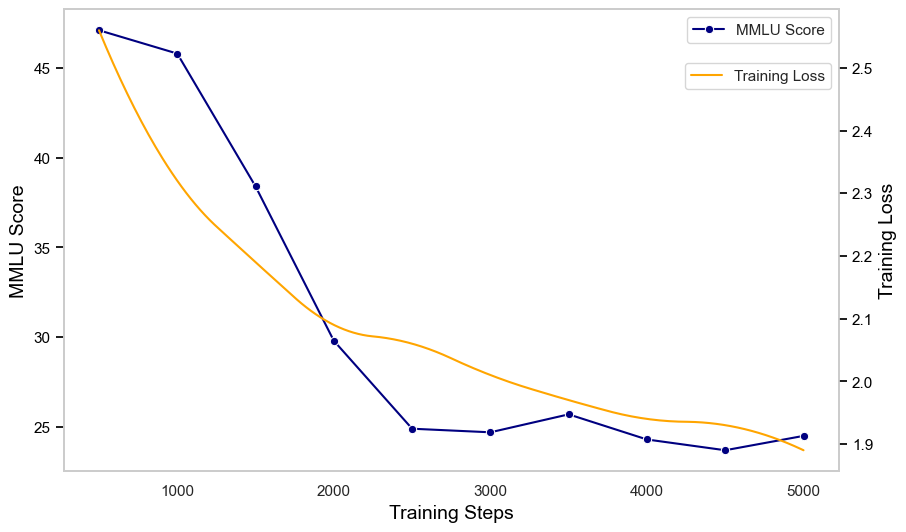} 
    \caption{\textbf{MMLU score and training loss over multimodal pretraining phase.} When directly training the language part of the VLM using the LAION dataset, the model's score on the pure text dataset MMLU rapidly decreases, dropping to 24.9 at 2500 steps.}
    \label{fig:mmlu}
  \end{minipage}\hfill
  \begin{minipage}{0.48\textwidth}
    \includegraphics[width=\linewidth]{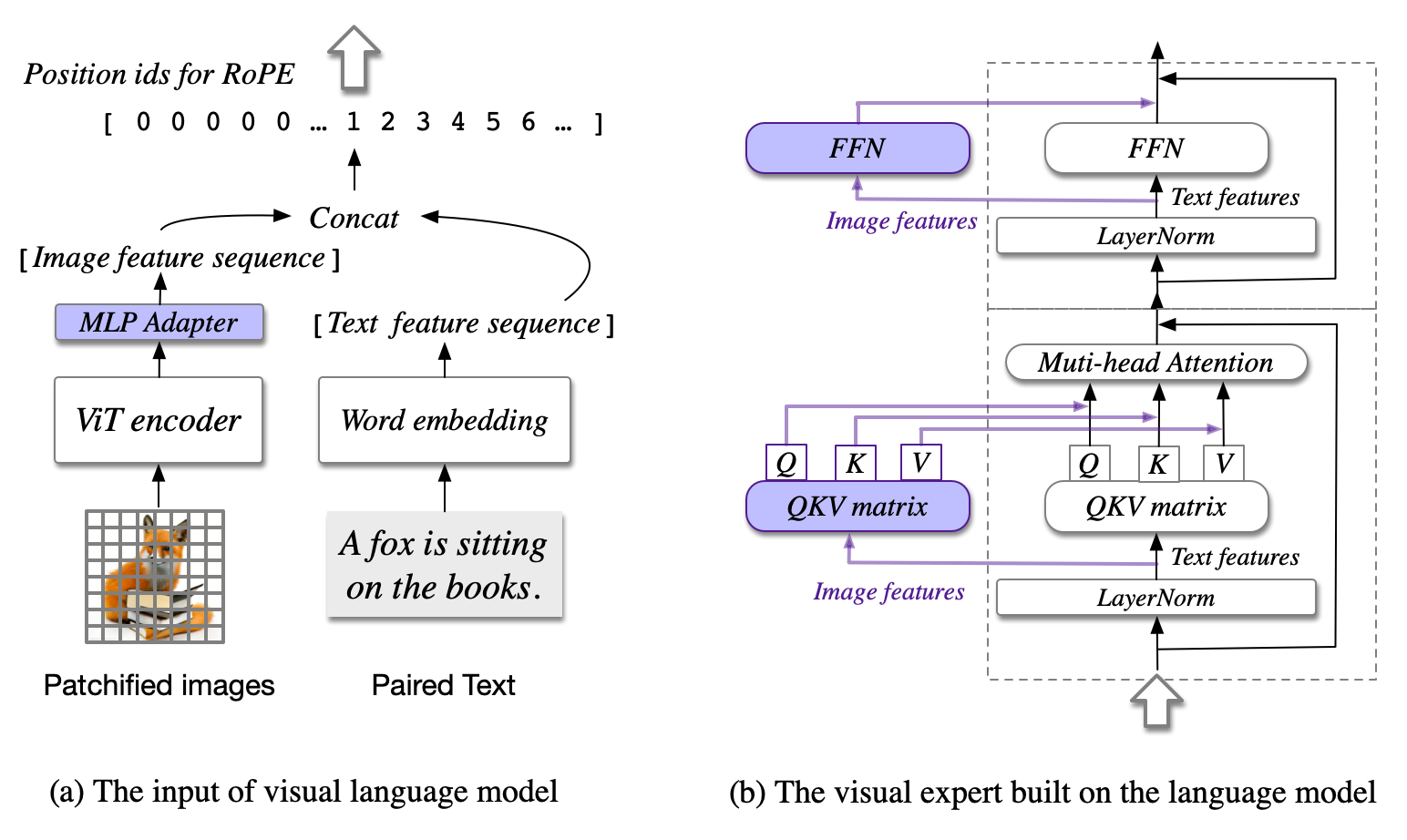} 
    \caption{\textbf{The architecture of CogVLM.} (a) The illustration about the input, where an image is processed by a pretrained ViT and mapped into the same space as the text features. (b) The Transformer block in the language model. The image features have a different QKV matrix and FFN. Only the \textcolor{violet}{purple} parts are trainable.}
    \label{fig:cogvlm_arch}
  \end{minipage}
\end{figure*}

Vision language models are versatile and powerful. Many vision and cross-modality tasks can be formulated as next token prediction, e.g., image captioning~\citep{agrawal2019nocaps}, visual question answering~\citep{antol2015vqa}, visual grounding~\citep{yu2016modeling} and even segmentation~\citep{chen2022generalist}. Useful abilities like in-context learning~\citep{tsimpoukelli2021multimodal, sun2023generative, alayrac2022flamingo} also emerge along with the improvement of downstream tasks when scaling up VLMs. 
However, to train a large language model is already non-trivial, and it is more challenging to train a VLM from scratch with the same NLP performance as well-trained pure language models like LLaMA2~\citep{touvron2023llama}. Therefore, it is natural to investigate how to train a VLM from an off-the-shelf pretrained language model.

The popular \emph{shallow alignment} methods represented by InstructBLIP~\citep{li2023blip} and MiniGPT-4~\citep{zhu2023minigpt} connect a frozen pretrained vision encoder and language model via a trainable Q-Former or a linear layer, mapping the image features into the input embedding space of the language model. 
This method converges rapidly, but its performance is noticeably inferior to that of LLaVA-1.5 with trainable language parameters, despite their model sizes and training datasets being almost identical.

The primary challenge in the performance of shallow alignment methods within VLMs can be attributed to the lack of deep fusion between visual and linguistic data. Shallow alignment methods struggle because they rely on `frozen' language model weights, which are intrinsically trained to process text tokens. This presents a significant mismatch issue, as visual features lack a direct equivalent in the textual input space. Consequently, when these visual features undergo multi-layer transformations, they tend to deviate from the expected input distribution of the deeper language model layers. This misalignment is particularly evident in tasks like image captioning, where the specificity of a task – such as writing style and caption length – can only be superficially encoded into visual features through shallow methods. 
% This superficial encoding not only weakens the consistency between the visual features and the actual content but also increases the risk of generating inaccurate or `hallucinated' content. 
% Addressing this issue calls for a more integrated approach where deeper and more nuanced connections between visual and textual elements are established during the model training process.
    
% A straightforward approach involves directly training large language models during the pre-training and supervised fine-tuning (SFT) phases to adapt them to multimodal inputs, which is adopted by PaLI~\citep{chen2022pali} and Qwen-VL~\citep{bai2023qwen}. However, this method can significantly impact the generalizability of large language models and might affect text-centered tasks, such as image-based poetry creation or introducing the background story of images. Typically, LLMs are pre-trained on vast corpora of pure text, and their data distribution differs markedly from that of image-text pair datasets like LAION and COYO. Continuing multimodal pre-training on this foundation almost inevitably leads to catastrophic forgetting. As illustrated in Figure 1, as the model increasingly fits the LAION dataset, there's a noticeable decline in its MMLU~\citep{hendrycks2020measuring} score, confirming our hypothesis. Similar phenomena have also been found in Palm-E and Flamingo: making the language model trainable during VLM pretraining will drop 87.3\% NLG performance for 8B language model.

A common strategy, as seen in PaLI~\citep{chen2022pali} and Qwen-VL~\citep{bai2023qwen}, involves direct training of LLM during the pre-training or supervised fine-tuning (SFT) phase. However, this approach can compromise the models' generalizability, particularly for tasks focused on textual outputs. Conventionally, LLMs are pretrained on extensive text-only datasets~\cite{raffel2020exploring}, leading to a significant divergence in data distribution when compared to image-text pair datasets like LAION~\citep{schuhmann2022laion} and COYO~\cite{kakaobrain2022coyo700m}. This shift often results in catastrophic forgetting, a phenomenon where the model's proficiency in its original domain deteriorates. This issue is evident in Figure~\ref{fig:mmlu}, which shows a marked decline in MMLU~\citep{hendrycks2020measuring} score as the model becomes more attuned to the LAION dataset, thus validating our hypothesis. This trend is not isolated; similar effects have been observed in models like PaLM-E~\citep{driess2023palm} and Flamingo~\citep{alayrac2022flamingo}. For instance, adapting an 8B parameter language model for VLM pretraining can lead to an 87.3\% reduction in natural language generation (NLG) performance~\citep{driess2023palm}.

The discussion above raises an important question: is it possible to retain the NLP capabilities of the large language model while adding top-notch visual understanding abilities to it?

CogVLM gives a ``\emph{yes}'' answer. 
% More detailed reasons for the performance degradation of p-tuning and shallow alignment include:
% \begin{enumerate}
%     \item The frozen weights in the language model are trained for text tokens.
%     Visual features do not have a perfect counterpart in the input text space. Therefore, after multi-layer transformations, the visual features might no longer match the input distribution of the weights in the deep layers.
%     \item During pretraining, the prior of the image captioning task, for example, the writing style and caption length, can only be encoded into the visual features in the shallow alignment methods. It weakens the consistency between visual features and content.
% \end{enumerate}
CogVLM instead adds a trainable \emph{visual expert} to the language model. In each layer, the image features in the sequence use a new QKV matrix and MLP layer with the text features. Visual expert doubles the number of parameters while keeping the FLOPs the same. Since all the parameters in the original language model are fixed, the behaviors are the same as in the original language model if the input sequence contains no image. This inspiration arises from the comparison between P-Tuning~\citep{liu2023gpt} and LoRA~\citep{hu2021lora} in efficient finetuning, where p-tuning learns a task prefix embedding in the input while LoRA adapts the model weights in each layer via a low-rank matrix. As a result, LoRA performs better and more stable. A similar phenomenon might also exist in VLM, because in the shallow alignment methods, the image features act like the prefix embedding in P-Tuning. 

Our contributions in this work are as follows:
\begin{itemize}
    \item We introduce the CogVLM model, which deeply integrates visual and linguistic features while retaining the full capabilities of a pretrained large language model. CogVLM-17B, trained from Vicuna-7B, achieves state-of-the-art across 17 classic cross-modal benchmarks.
    \item Through extensive ablation studies, we validated the effectiveness of our proposed visual expert module and the importance of deep fusion.
    We further delved into multiple critical factors in multimodal pertaining, including the scale of visual encoder, variants of attention mask, the most impactful parameters in VLMs, and the necessity of incorporating self-supervised image loss, etc. 
    % We hope that these investigations can offer substantial insights into the multimodal domain.

    \item We have made the weights of CogVLM and the dataset used in the SFT phase available to the public. We anticipate that the open sourcing of CogVLM will significantly contribute to the research and industrial application of visual understanding.
\end{itemize}

% \textbf{Our CogVLM-17B trained from Vicuna-7B achieves state-of-the-art or the second-best performance on 14 classic cross-modal benchmarks}, including 1) image captioning datasets: NoCaps, Flicker30k, COCO, 2) VQA datasets: VQAv2, OKVQA, TextVQA, OCRVQA, ScienceQA, 3) LMM benchmarks: 3) visual grounding datasets: RefCOCO, RefCOCO+, RefCOCOg, Visual7W, 4) multiple choice datasets: TDIUC, ScienceQA. 

% Since \textbf{most previous famous VLMs are close-source}, including Flamingo~\citep{alayrac2022flamingo}, SimVLM~\citep{wang2021simvlm}, Coca~\citep{yu2022coca}, BEIT-3(1.9B)~\citep{wang2022image}, GIT2~\citep{wang2022git}, PaLI~\citep{chen2022pali}, PaLI-X~\citep{chen2023pali}, \textbf{we anticipate that the open-sourcing of CogVLM will greatly help the research and industrial application of visual understanding. }

\section{Method}
\subsection{Architecture}

% \begin{figure}[]
%     \centering
%     \includegraphics[width=\linewidth]{figures/cogvlm.png}
%     \caption{The architecture of CogVLM. (a) The illustration about the input, where an image is processed by a pretrained ViT and mapped into the same space as the text features. (b) The Transformer block in the language model. The image features have a different QKV matrix and FFN. Only the \textcolor{violet}{purple} parts are trainable.}
%     \label{fig:cogvlm_arch}
% \end{figure}
CogVLM model comprises four fundamental components: a vision transformer (ViT) encoder, an MLP adapter, a pretrained large language model (GPT), and a visual expert module. Figure~\ref{fig:cogvlm_arch} shows an overview of the CogVLM architecture. The components' design and implementation details are provided below:

\textbf{ViT encoder}. We utilize pretrained EVA2-CLIP-E~\citep{sun2023eva} in CogVLM-17B. Note that the final layer of ViT encoder is removed because it specializes in aggregating the [CLS] features for contrastive learning.

\textbf{MLP adapter}. To map the output of ViT into the same space as the text features from word embedding, we use an MLP adapter, a two-layer MLP (SwiGLU~\citep{shazeer2020glu}). For implementation convenience, all image features share the same position id in the language model. 
    
\textbf{Pretrained large language model}. CogVLM's model design is compatible with any off-the-shelf GPT-style pretrained large language model. Specifically, CogVLM-17B adopts Vicuna1.5-7B~\citep{chiang2023vicuna} for further training. A causal mask is applied to all the attention operations, including the attention between image features.
% In the implementation of CogVLM-17B, we utilize pretrained EVA2-CLIP-E~\citep{sun2023eva} and Vicuna1.5-7B~\citep{chiang2023vicuna} as the ViT encoders and language model respectively. The final layer of ViT encoder is removed because it specializes in aggregating the [CLS] features for contrastive learning.
% The MLP adapter is a two-layer MLP (SwiGLU~\citep{shazeer2020glu}) to map the output of ViT into the same space as the text features from word embedding. All images features share the same position id in language model. Causal mask is applied to all the attention operations.

\textbf{Visual expert module}. 
% As the shallow alignment methods may cause insufficient model expressiveness and unsuitable image feature transformation, 
We add a visual expert module to each layer to enable deep visual-language feature alignment. Specifically, the visual expert module in each layer consists of a QKV matrix and an MLP in each layer. The shapes of the QKV matrix and MLP are identical to those in the pretrained language model and initialized from them. The motivation is that each attention head in the language model captures a certain aspect of semantic information, while a \emph{trainable} visual expert can transform the image features to align with the different heads, therefore enabling deep fusion.

Formally, suppose that the input hidden states of an attention layer are $X \in \mathbb{R}^{B\times H\times (L_I+L_T) \times D}$, where $B$ is the batch size, $L_I$ and $L_T$ are the lengths of image and text sequences, $H$ is the number of attention heads, and $D$ is the hidden size. In the attention with visual expert, $X$ is first split as image hidden states $X_I$ and text hidden states $X_T$, and the attention is computed as:
% The visual expert module consists of a QKV matrix and an MLP in each layer. The shapes of QKV matrix and MLP are identical to those in the pretrained language model, and initialized from them. We assume that each attention head in the language model capture a certain aspect of semantic information, while a \emph{trainable} visual expert can transform the image features to align with the different heads to enable deep fusion. Specifically, suppose that the input hidden states of an attention layer are $X \in \mathbb{R}^{B\times H\times (L_I+L_T) \times D}$, where $B$ is the batch size, $L_I$ and $L_T$ are the lengths of image and text sequences, $H$ is the number of attention heads, and $D$ is the hidden size. In the attention with visual expert, 
% $X$ is first split as $X_I$ and $X_T$ and the attention is computed as:
\begin{align}
    \text{Attention} & (X, W_I, W_T) = \text{softmax}(\frac{\text{Tril}(QK^T)}{\sqrt{D}})V,
\end{align}
% \begin{align}
%     Q = \text{concat}(X_IW_I^Q, X_TW_T^Q), 
% \end{align}
% \begin{align}
%     K &= \text{concat}(X_IW_I^K, X_TW_T^K), \\
% % \end{align}
% % \begin{align}
%     V &= \text{concat}(X_IW_I^V, X_TW_T^V),
% \end{align}
\begin{align}
    Q &= \text{concat}(X_IW_I^Q, X_TW_T^Q), \\
    K &= \text{concat}(X_IW_I^K, X_TW_T^K), \\
    V &= \text{concat}(X_IW_I^V, X_TW_T^V),
\end{align}
% \end{align}
where $W_I, W_T$ are the QKV matrices of the visual expert and original language model, and \text{Tril}$(\cdot)$ means lower-triangular mask.
The visual expert in FFN layers performs similarly,  
\begin{align}
    \text{FFN}(X) = \text{concat}(\text{FFN}_{I}(X_I),  \text{FFN}_{T}(X_T)),
\end{align}
where FFN$_I$ and FFN$_T$ are the FFN of the visual expert and original language model.

\textbf{Position embedding.} In the RoPE within LLM, we allow all visual tokens to share a single position id, as they already encapsulate positional information when inputted into the ViT. This approach mitigates the impact of remote attenuation between tokens in the LLM. Given that an image can occupy hundreds to thousands of tokens, and a typical input sequence is structured as \textit{`\textless image embed\textgreater ~query'}, using conventional positional encoding would result in excessively lengthy encoding sequences. Moreover, it would lead the query to focus more on the image sequences closer to it, namely the lower part of an image.

\subsection{Pretraining}
\textbf{Data.} The image-text pairs for pretraining are all publicly available, including LAION-2B and COYO-700M. After removing the broken URLs, NSFW images, images with noisy captions, images with political bias and images with an aspect ratio $>6$ or $<1/6$, about 1.5B images are left for pretraining.

We also crafted a visual grounding dataset of 40M images. Each noun in the image caption is associated with bounding boxes to indicate the positions in the image. The construction process basically follows \cite{peng2023kosmos}, which extracts nouns via spaCy~\citep{honnibal2015improved} and predicts the bounding boxes using GLIPv2~\citep{zhang2022glipv2}. The image-text pairs are sampled from LAION-115M, a subset of LAION-400M filtered by \cite{li2023blip}. We filter and retain a subset of 40 million images to ensure that over 75\% of images contain at least two bounding boxes.
% This dataset is only used for training CogVLM-17B-chat model, and not used for evaluating downstream tasks, e.g. RefCOCO, for fair comparison.

% The LAION40M dataset is derived from the foundation of LAION115M~\cite{li2023blip}, which is an image captioning dataset encompassing 115 million image-text pairs. We enhanced this foundational dataset by incorporating visual grounding information. The methodology employed for this grounding procedure involves the use of GLIPv2~\cite{zhang2022glipv2}. When provided with an image and a noun phrase, GLIPv2 can identify and annotate the corresponding region(s) in the image. To harness this capability, we utilized the Spacy toolkit to extract noun phrases from the textual descriptions within LAION115M. Subsequently, these noun phrases were fed into GLIPv2 to generate the corresponding bounding boxes. This procedure resulted in the grounding of all 115 million data points. We further refined the dataset, ensuring images typically contained multiple bounding boxes. However, we did retain a small fraction of images with only a single box, preserving them with a 0.25\% probability. This filtration process yielded a consolidated dataset of 40 million entries, which we term as LAION40M.

\textbf{Training.} 
The first stage of pretraining is for \emph{image captioning loss}, i.e. next token prediction in the text part. We train the CogVLM-17B model on the 1.5B image-text pairs introduced above for 120,000 iterations with a batch size of 8,192.
The second stage of pretraining is a mixture of image captioning and Referring Expression Comprehension (REC). REC is a task to predict the bounding box in the image given the text description of an object, which is trained in the form of VQA, i.e., \textit{Question: Where is the \emph{object}?} and \textit{Answer: $[[x_0, y_0, x_1, y_1]]$}. Both $x$ and $y$ coordinates range from $000$ to $999$, meaning the normalized position in the image.
We only consider the loss of the next token prediction in the ``Answer'' part. 
We pretrain the second stage for 60,000 iterations with a batch size of 1,024 on the text-image pairs and visual grounding datasets introduced above. During the final 30,000 iterations, we change the input resolution from $224\times 224$ to $490\times 490$. The total number of trainable parameters is 6.5B. 

\begin{table*}[h]
\centering
\caption{\textbf{Performance on Image Captioning benchmarks.} All tasks use CIDEr as the evaluation metric. OOD refers to out-of-domain test set. Karp. refers to the Karpathy test split.
% VinVL~\citep{zhang2021vinvl}, SimVLM~\citep{wang2021simvlm}, CoCa~\citep{yu2022coca}, LEMON~\citep{hu2022scaling}, Flamingo~\citep{alayrac2022flamingo}, Prismer~\citep{liu2023prismer}, BLIP-2~\citep{li2023blip}, InstructBLIP~\citep{dai2023instructblip}, InstructBLIP~\citep{dai2023instructblip}, UniversalCap~\citep{cornia2021universal}, GIT~\citep{wang2022git}, GIT2~\citep{wang2022git}, PaLI-17B~\citep{chen2022pali}, Pali-X-55B~\citep{chen2023pali}
}
\label{caption}
\vskip 0.1in
  \resizebox{\textwidth}{!}{%
  \centering
  \renewcommand{\arraystretch}{1.1}
  \setlength{\tabcolsep}{10pt}
  {\small
    \begin{tabular}
    % {lc{0.8cm}c{0.8cm}c{0.8cm}c{0.8cm}c{0.8cm}c{0.8cm}c{0.8cm}c{1.1cm}}
    {lcccccccc}
    \toprule
    \multirow{2}{*}{\textbf{Method}} & \multirow{2}{*}{\begin{minipage}{1cm}\textbf{Train Data}\end{minipage}} & \multicolumn{2}{c}{\textbf{NoCaps val}} & \multicolumn{2}{c}{\textbf{NoCaps test}} & \multirow{1}{*}{\textbf{Flickr}} & \multirow{1}{*}{\textbf{COCO}} & \multirow{1}{*}{\textbf{TextCaps}} \\
    \cmidrule(r){3-4} \cmidrule(r){5-6} \cmidrule(r){7-7} \cmidrule(r){8-8} \cmidrule(r){9-9}
    & & OOD & overall & OOD & overall & Karp. & Karp. & test\\
        \midrule
        Human & - & 95.7 & 87.1 & 91.6 & 85.3 & - & - & 125.1 \\
        VinVL~\citep{zhang2021vinvl} & 8.9M & 83.8 & 94.3 & 78.0 & 92.5 & - & 130.8 & - \\
        SimVLM~\citep{wang2021simvlm} & 1.8B & 115.2 & 112.2 & 109.5 & 110.3 & - & 143.3 & - \\
        CoCa~\citep{yu2022coca} & 4.8B & - & 122.4 & - & 120.6 & - & 143.6 & - \\
        LEMON~\citep{hu2022scaling} & 2B & 120.2 & 117.3 & 110.1 & 114.3 & - & 139.1 & - \\
        Flamingo~\citep{alayrac2022flamingo} & 2.3B & - & - & - & - & 67.2 & 138.1 & - \\
        Prismer~\citep{liu2023prismer} & 12.7M & 113.5 & 112.9 & - & 110.8 & - & 136.5 & - \\
        BLIP-2~\citep{li2023blip} & 129M & 124.8 & 121.6 & - & - & - & 144.5 & - \\
        InstructBLIP~\citep{dai2023instructblip} & 129M & - & 123.1 & - & - & 82.4 & - & - \\
        UniversalCap~\citep{cornia2021universal} & 35M & 123.4 & 122.1 & 114.3 & 119.3 & - & 143.4 & - \\
        GIT~\citep{wang2022git} & 0.8B & 127.1 & 125.5 & 122.0 & 123.4 & 49.6 & 144.8 & 138.2 \\
        GIT2~\citep{wang2022git} & 12.9B & \underline{130.6} & \underline{126.9} & \underline{122.3} & \underline{124.8} & 50.7 & 145.0 & \underline{145.0} \\
        Qwen-VL~\citep{bai2023qwen} & 1.4B & - & 121.4 & - & - & 85.8 & - & - \\
        PaLI-17B~\citep{chen2022pali} & 1.6B & - & 127.0 & - & 124.4 & - & \underline{149.1} & 135.4 \\
        PaLI-X-55B~\citep{chen2023pali} & - & - & 126.3 & - & 124.3 & - & \textbf{149.2} & \textbf{147.0} \\
        \midrule
        CogVLM (ours) & 1.5B & \textbf{132.6} & \textbf{128.3} & \textbf{128.0} & \textbf{126.4} & \textbf{94.9} & 148.7 & 144.9 \\
%     Human & 91.6 & 14.2 & 85.3 & 14.6 & 87.1 & 14.2 & - \\
%     VinVL & 78.0 & 11.5 & 92.5 & 13.1 & 94.3 & 13.1 & - \\
    \bottomrule
    \end{tabular}
  } 
  }
    % \vspace{2mm}
% \caption{Performance comparison of state-of-the-art image captioning methods on various datasets. All tasks use CIDEr as the evaluation metric. VinVL~\citep{zhang2021vinvl}, SimVLM~\citep{wang2021simvlm}, CoCa~\citep{yu2022coca}, LEMON~\citep{hu2022scaling}, Flamingo~\citep{alayrac2022flamingo}, Prismer~\citep{liu2023prismer}, BLIP-2~\citep{li2023blip}, InstructBLIP~\citep{dai2023instructblip}, InstructBLIP~\citep{dai2023instructblip}, UniversalCap~\citep{cornia2021universal}, GIT~\citep{wang2022git}, GIT2~\citep{wang2022git}, PaLI-17B~\citep{chen2022pali}, Pali-X-55B~\citep{chen2023pali}
% }
  % \label{caption}
\vskip -0.1in
\end{table*}
% >{\centering}p{1.2cm}
\subsection{Alignment}
\label{sec:alignment}
In the instruction alignment phase, we trained two generalist models: CogVLM-Chat and CogVLM-Grounding. CogVLM-Chat accepts natural language inputs and outputs, while CogVLM-Grounding accepts inputs and outputs with bounding boxes.

\textbf{CogVLM-Chat.} 
In our study, we integrated data from a variety of open-source visual question-answering datasets, including VQAv2~\citep{antol2015vqa}, OKVQA~\citep{marino2019ok}, TextVQA~\citep{singh2019towards}, OCRVQA~\cite{mishra2019ocr}, ScienceQA~\citep{lu2022learn}, as well as datasets formatted as multi-turn dialogues such as LLaVA-Instruct~\citep{liu2023visual}, LRV-Instruction~\citep{liu2023aligning}, LLaVAR~\cite{zhang2023llavar}. We then conducted unified instruction-supervised fine-tuning (SFT) across these diverse datasets. The integrity and quality of SFT data are crucial; notably, the LLaVA-Instruct dataset, initially generated through a language-only GPT-4 pipeline, contained certain inaccuracies. We meticulously corrected these errors through manual inspection and annotation to ensure data quality. 

VQA datasets typically feature concise, often one-word answers, contrasting with the dialogue datasets that provide detailed responses with extensive reasoning. To accommodate this variability, we employed prompts formatted as \textit{Question: Short answer:} for concise responses and \textit{Question: Answer:} for extended discourse in the SFT phase. 

During training, the model underwent 6000 iterations with a learning rate of 1e-5 and a batch size of 1024. To enhance and ensure the stability of the training, we activated the visual encoder's parameters and adjusted its learning rate to be one-tenth of that used for the remaining training parameters.

\textbf{CogVLM-Grounding.} 
In order to endow our model with consistent, interactive visual grounding capabilities, we collect a high-quality dataset covering 4 types of grounding data: (1) \textbf{Grounded Captioning (GC)} - image captioning datasets where each noun phrase within the caption is followed by the corresponding referential bounding boxes; (2) \textbf{Referring Expression Generation (REG)} - image-oriented datasets that each bounding box in the image is annotated with a descriptive textual expression that accurately characterizes and refers to the content within the specific region; (3) \textbf{Referring Expression Comprehension (REC)} - text-oriented datasets that each textual description is annotated with multiple referential links associating the phrases with corresponding boxes; (4) \textbf{Grounded Visual Question Answering (GroundedVQA)} - VQA-style datasets where the questions may contain region references in a given image.
The sources of grounding data are all publicly available, including Flickr30K Entities~\citep{plummer2015flickr30k}, RefCOCO~\citep{kazemzadeh2014referitgame,mao2016generation,yu2016modeling}, Visual7W~\citep{zhu2016visual7w}, VisualGenome~\citep{krishna2017visual} and Grounded CoT-VQA~\citep{chen2023shikra}.
$[box]$ in this section is in the format of $[[x_0, y_0, x_1, y_1]]$. 

It is noteworthy that the curated datasets exhibit a versatility of visual grounding capabilities, and many datasets can be adapted and repurposed across different tasks.
For instance, grounded captioning datasets can be reformulated to suit REG and REC tasks. Taking the example of \textit{``A man $[box_1]$ and a woman $[box_2]$ are walking together.''}, this can be reframed into question answering pairs like \textit{(``Describe this region $[box_2]$.'', ``A woman.'')} and \textit{(``Where is the man?'', ``$[box_1]$'')}. Similarly, REC datasets can be translated into REG tasks by switching the input and output, and vice versa. However, certain conversions might lead to ambiguities. For example, when presented with the isolated query ``Where is another man?'' from the caption ``A man $[box_1]$ is running, while another man $[box_2]$ is looking.'', the distinction between $[box_1]$ and $[box_2]$ becomes unclear, potentially leading to errors.

\section{Experiments}

% \dm{template\\
% \textbf{Datasets. }One sentence to introduce the task, and then introduce the datasets. \begin{itemize}
%     \item \textbf{DS1}. xxx
%     \item \textbf{DS2}. xxx
% \end{itemize}
% \textbf{Settings. } How to train and the metrics. \\
% \textbf{Results. } Table and analysis.
% }

% To rigorously validate the superior performance and robust generalization of our model across multimodal tasks, we conducted evaluations on an array of established benchmarks. These evaluations span three primary domains:
To rigorously validate the superior performance and robust generalization of our base model, we conduct quantitative evaluations on an array of multi-modal benchmarks. These benchmarks can be categorized into three broad areas covering a comprehensive range of measurement\footnote{Detailed summary of all benchmarks and corresponding metrics are available at Appendix~\ref{subsec:details_of_eval_benchmarks}.}:
% \begin{itemize}
%     \item \textbf{Image Captioning}. We utilized prominent datasets including NoCaps~\citep{agrawal2019nocaps}, COCO~\citep{lin2014microsoft}, Flickr30K~\citep{plummer2015flickr30k}, and TextCaps~\citep{sidorov2020textcaps}.
%     \item \textbf{Visual Question Answering}. Our assessment covered diverse datasets such as VQAv2~\citep{antol2015vqa}, OKVQA~\citep{marino2019ok}, TextVQA~\citep{singh2019towards}, VizWiz-VQA~\citep{gurari2018vizwiz}, OCRVQA~\citep{mishra2019ocr}, ScienceQA~\citep{lu2022learn}, and TDIUC~\citep{shrestha2019answer}.
%     \item \textbf{Visual Grounding}. We incorporated datasets like Visual7w~\citep{zhu2016visual7w}, RefCOCO~\citep{liu2017referring}, RefCOCO+, and RefCOCOg to ensure a comprehensive evaluation.
% \end{itemize}
\begin{itemize}
    \item \textbf{Image Captioning}. The main purpose of these tasks is to generate textual captions summarizing the major content of a given image. We utilize prominent datasets including NoCaps~\citep{agrawal2019nocaps}, COCO~\citep{lin2014microsoft}, Flickr30K~\citep{plummer2015flickr30k}, and TextCaps~\citep{sidorov2020textcaps} for evaluation.
    \item \textbf{Visual Question Answering}. The VQA tasks require models to answer questions that may focus on distinct visual contents based on the given image. Our assessment covers diverse datasets, including VQAv2~\citep{antol2015vqa}, OKVQA~\citep{marino2019ok}, TextVQA~\citep{singh2019towards}, OCRVQA~\cite{mishra2019ocr} and ScienceQA~\citep{lu2022learn}.
    \item \textbf{LVLM Benchmarks}. LVLM benchmarks are primarily employed to assess the advanced capabilities of large multimodal models, such as object recognition and localization, OCR, visual description, and visual knowledge reasoning. We conduct multidimensional evaluations of the models on datasets including MM-Vet~\cite{yu2023mm}, MMBench~\cite{liu2023mmbench}, SEED-Bench~\cite{li2023seed}, LLaVA-Bench~\cite{liu2023visual}, POPE~\cite{li2023evaluating}, MMMU~\cite{yue2023mmmu} and MathVista~\cite{lu2023mathvista}.
    \item \textbf{Visual Grounding}. Visual grounding involves a set of tasks that establish referential links between textual mentions in a sentence and specific regions in an image. We evaluate our model on the typical datasets, including Visual7w~\citep{zhu2016visual7w}, RefCOCO~\citep{liu2017referring}, RefCOCO+, and RefCOCOg to ensure completeness.
\end{itemize}

\subsection{Image Captioning}

We evaluate the image captioning capability of our pretrained base model on the aforementioned four benchmarks. In a zero-shot evaluation on the Nocaps and Flickr datasets, we assess the precision of our model in describing long-tail visual concepts. Additionally, we present results from finetuning on the COCO and TextCaps datasets.

\begin{table*}[]
 % \caption{Comparison with generalist models.}
 \caption{\textbf{Generalist performance on VQA and LVLM benchmarks.} * donates the dataset has been trained during SFT stage. We compared with the latest state-of-the-art generalist models, including MiniGPT-4~\citep{zhu2023minigpt}, IDEFICS-Instruct~\citep{laurencon2023obelics}, OpenFlamingo~\cite{awadalla2023openflamingo}, DreamLLM~\cite{dong2023dreamllm}, InstructBLIP~\cite{dai2023instructblip}, Fuyu~\cite{fuyu-8b}, Qwen-VL~\cite{bai2023qwen}, LLaVA-1.5~\cite{liu2023improved}, mPLUG-Owl2~\cite{ye2023mplug}, SPHINX~\cite{lin2023sphinx}, Emu2~\cite{sun2023generative}.}
 \vskip 0.1in
  \resizebox{\textwidth}{!}{%
  \centering
  \renewcommand{\arraystretch}{1.15}
  \setlength{\tabcolsep}{1pt}
  \small
  {
    \begin{tabular}{l|c|ccccc|ccccccc}
    \toprule
    % \multirow{2}{*}{\textbf{Method}} & \multirow{2}{*}{\textbf{LLM}} & \multicolumn{1}{c}{\textbf{VQAv2}} & \multicolumn{1}{c}{OKVQA} & \multicolumn{1}{c}{TextVQA} & \multicolumn{1}{c}{\textbf{ScienceQA}} & \multicolumn{1}{c}{\textbf{MM-Vet}} & \multicolumn{1}{c}{\textbf{SEED}} & \multicolumn{1}{c}{\textbf{MMBench}} & \multicolumn{1}{c}{\textbf{POPE}}  & \multicolumn{1}{c}{\textbf{MMMU}}  & \multicolumn{1}{c}{\textbf{MathVista}}\\
    \multirow{2}{*}{\textbf{Method}} & \multirow{2}{*}{\textbf{LLM}} & \multicolumn{5}{c}{\textbf{VQA}}  & \multicolumn{6}{c}{\textbf{LVLM-Benchmark}} \\
    & &VQAv2 & OKVQA & TextVQA & OCRVQA & ScienceQA & MM-Vet & SEED & MMBench & LLaVA& POPE & MMMU & MathVista\\
    \midrule
    MiniGPT-4 & Vicuna-7B & - & - & 0.6 & - & 39.6 & 22.1 & 47.4 & 23.0 & 45.1 & - & - & 23.1\\
    IDEFICS-Instruct & LLaMA-65B & 37.4 & 36.9 & 32.9 & - & 61.8 & 39.7 & 53.2 & 54.5 & 56.9& - & - & 26.2 \\ %39.7
    OpenFlamingo & MPT-7B & 53.0 & 38.3 & 28.3 & - & 44.8 & 24.8 & 42.7 & 5.7 & 34.2& - & 26.3 & 18.6 \\
    DreamLLM & Vicuna-7B & 56.6 & 44.3 & 34.9 & - & - & 35.9 & - & 49.9 & - & - & - & -\\
    InstructBLIP & Vicuna-7B & - & - & 50.1 & - & 60.5 & 26.2 & 58.8 & 33.9 & 59.8 & 53.8 & - & 25.3\\
    Fuyu & Fuyu-8B & 74.2* & 60.6* & - & - & - & - & - & - & - & - & 27.4 & - \\
    Qwen-VL-Chat & Qwen-7B  & 78.2* & 56.6* & 61.5* & \underline{70.5*} & 68.8 & - & 65.4 & 61.8 & 67.7 & - & 32.9 & \underline{33.8}\\
    LLaVA-1.5 & Vicuna-7B & 78.5* & - & 58.2* & - & 66.8 & 30.5 & 58.6 &  64.3 & 60.7& 85.9 & - & 23.6\\
    mPLUG-Owl2 & LLaMA2-7B & 79.4* & 57.7* & 58.2* & - & 68.7 & 36.2 & 64.1 & 64.5 & 25.0 & 86.2 & 32.1 & 25.3\\
    Unified-IO2 & UIO-2XXL & 79.4* & 55.5* & - & - & \underline{86.2*} & - & 65.6 & \underline{71.5} & - & \underline{87.7} & - & -\\
    LLaVA-1.5 & Vicuna-13B & 80.0* & - & 61.3* & - & 71.6 & 35.4 & 61.6 &  67.7 & 64.6 & 85.9 & 33.6 & 26.1\\
    SPHINX-2k & LLaMA2 13B & 80.7* & 62.6* & 61.2* &67.8* & 70.6 & 40.2 & \underline{71.6} & 65.9 & -& 87.2 & 32.9 & 27.8\\
    Emu2-Chat & LLaMA-33B & \textbf{84.9*} & \underline{64.8*} & \underline{66.6*} & -  & - & \underline{48.5} & 62.8 & 63.6 & 56.4 & - & \underline{34.1} & -\\
    \midrule
    CogVLM-Chat & Vicuna-7B & \underline{82.3*} & \textbf{64.8*} & \textbf{70.4*} & \textbf{73.8*} & \textbf{91.2*} & \textbf{51.1} & \textbf{72.5} & \textbf{77.6} & \textbf{77.8} & \textbf{87.9} & \textbf{41.1} & \textbf{34.5} \\

    \bottomrule
    \end{tabular}
    \label{tab:generalist}
  }
 }
\vskip -0.1in
\end{table*}
The detailed performance is shown in Table~\ref{caption}.  Overall, our model achieves the SOTA or compatible performance across the board. Specifically, on the NoCaps benchmark, our base model outperforms the previous best method, GIT2, across four splits with a maximum of $5.7$ points in the out-domain set while only consuming 10\% of the pretraining data (1.5B vs 12.9B). On the Flickr benchmark, our model achieves a SOTA score of $94.9$ surpassing the concurrently released Qwen-VL model by $9.1$ points. These results demonstrate the remarkable capability and robustness of our pretrained model on the image captioning task.
We also evaluate our model on the COCO~\citep{lin2014microsoft} and TextCaps, where the latter is specifically designed to integrate the textual information of the given image into captions.
Though training without the dedicated OCR data, encouragingly, our base model reveals a significant text-reading ability and obtains a competitive performance with PaLI-X-55B, and outperforms the previous best model of the same scale, PaLI-17B, by $9.1$ points score.

\begin{table*}[h]
\centering
\caption{Results on Referring Expression Comprehension and Grounded Visual Question Answering.}
\vskip 0.1in
  \resizebox{\textwidth}{!}{%
% \begin{tabular}{p{1.5cm}|l|cp{0.9cm}p{0.9cm}p{0.9cm}p{0.9cm}p{0.9cm}p{0.9cm}p{0.9cm}c}
\begin{tabular}{p{1.5cm}|l|ccccccccc}
\toprule
\multirow{2}{*}{\quad \textbf{Type}} & \multirow{2}{*}{\quad\quad\quad\quad\quad\, \textbf{Model}} & \multicolumn{3}{c}{\textbf{RefCOCO}} & \multicolumn{3}{c}{\textbf{RefCOCO+}} & \multicolumn{2}{c}{\textbf{RefCOCOg}} & \textbf{Visual7W} \\
\cmidrule(r){3-5} \cmidrule(r){6-8} \cmidrule(r){9-10} \cmidrule(r){11-11}
 &  & val & test-A & test-B & val & test-A & test-B & val & test & test \\ \midrule
\multirow{7}{*}{\textit{Generalist}} & OFA-L*~\citep{wang2022ofa} & 79.96 & 83.67 & 76.39 & 68.29 & 76.00 & 61.75 & 67.57 & 67.58 & -\\
 & VisionLLM-H~\citep{wang2023visionllm} &  - & 86.70 & - & - & - & - & - & - & -\\
 & Shikra-7B~\citep{chen2023shikra} & 87.01 & 90.61 & 80.24 & 81.60 & 87.36 & 72.12 & 82.27 & 82.19 & - \\
 & Shikra-13B~\citep{chen2023shikra} & 87.83 & 91.11 & 81.81 & 82.89 & 87.79 & 74.41 & 82.64 & 83.16 & 85.33\\
 & Qwen-VL~\citep{bai2023qwen} & 89.36 & 92.26 & 85.34 & 83.12 & 88.25 & 77.21 & 85.58 & 85.48 & -\\
 & Ferret-13B~\citep{you2023ferret} & 89.48 & 92.41 &84.36 &82.81 &88.14 &75.17 &85.83 &86.34 & -\\
 % & Qwen-VL-7B-Chat & 88.55 & 92.27 & 84.51 & 82.82 & 88.59 & 76.79 & 85.96 & 86.32 & -\\ 
  % & \textbf{CogVLM (chat)} & 90.54 & 92.96 & 86.26 & 84.33 & 89.68 & 77.21 & 87.93 & 88.42 & 89.38 \\
 & \textbf{CogVLM-Grounding} & \textbf{92.76} & \textbf{94.75} & \textbf{88.99} & \textbf{88.68} & \textbf{92.91} & \textbf{83.39} & \textbf{89.75} & \textbf{90.79} & \textbf{91.05} \\ \midrule
\multirow{4}{*}{\textit{Specialist}} & G-DINO-L~\cite{liu2023grounding} & 90.56 & 93.19 & 88.24 & 82.75 & 88.95 & 75.92 & 86.13 & 87.02 & -\\
 & UNINEXT-H~\citep{lin2023uninext} & 92.64 & 94.33 & 91.46 & 85.24 & 89.63 & 79.79 & 88.73 & 89.37 & -\\
 & ONE-PEACE~\citep{wang2023one} & 92.58 & 94.18 & 89.26 & 88.77 & 92.21 & 83.23 & 89.22 & 89.27 & -\\
 \bottomrule
 % \vspace*{-1cm}
\end{tabular}}
\label{tab:grounding}
\vskip -0.1in
\end{table*}

\subsection{Visual Question Answering}
% Visual Question Answering is a task of validating general multi-modal capabilities of models, which requires a mastery of skills including vision-language understanding and commonsense reasoning. We evaluate our model on 4 VQA benchmarks: VQAv2, OKVQA, TextVQA, ScienceQA, covering a wide range of visual scenes.
As illustrated in Table~\ref{tab:generalist}, our CogVLM model demonstrates outstanding performance and a significant lead over models of similar parameter scale across a variety of tasks, including daily-life image question-answering dataset VQAv2, text-intensive image question-answering datasets such as TextVQA and OCRVQA, and knowledge-demanding datasets like OKVQA and ScienceQA. This success showcases the model’s robust generalization capabilities and potential across diverse domains.

\subsection{LVLM Benchmarks}
Our findings, detailed in Table~\ref{tab:generalist}, demonstrate that CogVLM achieved state-of-the-art results in all 7 LVLM-benchmarks, markedly surpassing all other models. It also outperformed multimodal models that utilized larger language models, such as LLava1.5 with Vicuna-13B and Emu-2 with LLAMA-33B, leading by 15.7 and 2.6 points on MM-vet, 9.9 and 14.0 points on MMBench, respectively. Compared to IDEFICS-Instruct trained on LLaMA-65B, CogVLM's scores exceeded by 19.3, 23.1, and 20.9 points on Seed-Bench, MMBench, and LLaVA-Bench, respectively. Furthermore, CogVLM achieved a score of 41.1 on the MMMU dataset, and also scored 87.9 on the hallucination assessment dataset POPE, along with 35.2 on the multimodal mathematical reasoning benchmark MathVista. These impressive results not only showcase its robust reasoning abilities and multi-task generalization capabilities but also clearly demonstrate that CogVLM is significantly outpacing other models in these domains. Notably, shallow fusion models such as InstructBLIP and MiniGPT-4 underperformed across most benchmarks, despite InstructBLIP's extensive training on instructional data, underscoring the necessity of deep fusion for enhanced performance.

\begin{table*}[h]
% \vspace*{-0.1cm}

\caption{{Ablation studies for various components and training settings. \textit{VE} refers to visual expert.} 
}
\vskip 0.1in
\resizebox{\textwidth}{!}{%
  {\small
\begin{tabular}{cc|cc|c|ccccc}
\toprule
& \multirow{2}{*}{\textbf{Ablated Aspects}}   & \multirow{2}{*}{\textbf{Original Setting}}  & \multirow{2}{*}{\textbf{Ablated Setting}} & \small{\textbf{Trainable}}& \textbf{COCO} & \textbf{NoCaps} & \textbf{OKVQA}  & \textbf{TextVQA} & \textbf{VQAv2}  \\
& & & & \small{\textbf{params}} & CIDEr$\uparrow$  &  CIDEr$\uparrow$ & top1$\uparrow$     & top1$\uparrow$   & top1$\uparrow$  \\ \midrule

& \multirow{4}{*}{Tuned parameters} & \multirow{4}{*}{\begin{minipage}{2.8cm}
    \begin{center}
        \textit{VE-full} every layer \\ + \\ MLP Adapter
    \end{center}
\end{minipage}} & \multicolumn{1}{c|}{MLP Adapter} & 140M & 131.2	& 111.5	& 55.1	& 40.7	& 73.8\\
% & \multirow{4}{*}{Visual Expert (\textit{VE})} & \multirow{4}{*}{\textit{VE (Attn+FFN)} every layer} & \multicolumn{1}{l|}{\textit{No VE}, tune Adapter} & 140M & 131.2	& 111.5	& 55.1	& 40.7	& 73.8\\
& &  & \multicolumn{1}{c|}{LLM+MLP Adapter} & 6.9B & 140.3	& 118.5	& 56.8	& 44.7	& 78.9\\
& & & \multicolumn{1}{c|}{\textit{VE-full} every 4th layer} & 1.7B & 138.7	& 117.4	& 58.9	& 44.1 & 77.6\\
& & & \multicolumn{1}{c|}{\textit{VE-FFN} every layer} & 4.4B & 140.0	& 118.7	& 58.2	& 45.1 & 78.6\\
\midrule
% & Pretraining data & CLAY & LAION2B + COYO700M & 6.6B\\
% \midrule
& Init method & From LLM & Random init & 6.6B & 138.0	& 117.9	& 55.9	& 44.0 & 79.1\\
\midrule
& Visual attention mask & Causal mask & Full mask & 6.6B & 141.0 & 117.2 &	57.4 &	45.1 &	79.6 \\
\midrule
& Image SSL loss & \XSolidBrush & \Checkmark(clip feature) & 6.6B & 142.9 & 119.8 & 58.7 & \textbf{45.9} & 79.7\\
\midrule
& Visual encoder & EVA2-E & EVA2-L & 6.6B & 141.4 & \textbf{122.5} & 59.2 & 42.8 & 79.0\\
\midrule
% &Freezing ViT & \Checkmark & \XSolidBrush & 10.9B & 143.8 & 123.2 &	60.1 & 48.1	& 80.1 \\
% \midrule
% &Unify position & \Checkmark & \XSolidBrush & 6.6B\\
% \midrule
&EMA & \Checkmark & \XSolidBrush & 6.6B & \textbf{143.1} & 119.2 & 57.1 & 43.8 & 79.4\\
\midrule
% &\multicolumn{3}{c|}{\textbf{ model}}  & 6.6B & 142.8	& 120.1	& 59.3	& 45.3	& 80.0 \\ 
& \textit{CogVLM (ours)} & --- & --- & 6.6B & 142.8	& 120.1	& \textbf{59.3}	& 45.3	& \textbf{80.0} \\ 
\midrule
% ... (rest of the table remains unchanged)
\end{tabular}%
}}
% \vspace*{-0.1cm}
\label{tab:ablation-table-no-classif}
\vskip -0.1in
\end{table*}

\subsection{Visual Grounding}
Table~\ref{tab:grounding} shows the result on the standard visual grounding benchmarks.
We find that our generalist model achieves state-of-the-art performance across the board, with a significant advantage over the previous or concurrent models. As shown in the bottom part of Table~\ref{tab:grounding}, our model even surpasses models that are specifically trained for individual tasks, achieving SOTA performance on 5 of 9 splits. For instance, in the RefCOCO val subset, our model attains a score of 92.76, surpassing UNINEXT-H's 92.64; in the RefCOCO+ test-A subset, it scores 92.91, exceeding ONE-PEACE's 92.21; and in the RefCOCOg test subset, it achieves 90.79, outperforming UNINEXT-H's 89.27. These results suggest a remarkable visual grounding capability of our model incorporating our training paradigm.

\subsection{Ablation Study}
To understand the impact of various components and settings on our model's performance, we conduct an extensive ablation study for 6,000 iterations and a batch size of 8,192. Table~\ref{tab:ablation-table-no-classif} summarizes the results about the following aspects:

\textbf{Model structure and tuned parameters}. To investigate the effectiveness of CogVLM's model, we conduct ablation studies on several structure variants and tuning strategies, including: 1) tuning only the MLP Adapter layer; 2) tuning all LLM parameters and the Adapter without adding visual expert; 3) only adding visual expert at every 4th LLM layer; and 4) only add visual expert to FFNs at all layers.

From the results, we can see that shallow vision-language alignment, i.e. only tuning the adapter layer (similar to the method used in BLIP-2), results in a significantly inferior performance. Also, the performance of training the visual expert is higher than that of training the LLM, especially on the datasets that require external knowledge, even though the training parameters are roughly the same. We also compare with other variants of adding visual expert, including a. inserting an expert module every 4 layers and b. removing the attention part from the expert. Both of them result in a certain degree of performance decline, but within an acceptable range, which provides some guidance for balancing computational overhead and model performance.
\textbf{Initialization Method}. As for visual expert's initialization method, we compare initialization with weights from LLM to random initialization.
% investigate the effectiveness of initializing visual expert weights from LLM. 
Our results across various datasets demonstrate that initialization with LLM's weights consistently achieves superior performance. This indicates that the transformer architecture pre-trained on language data possesses a certain capability to process visual tokens. Moreover, it can serve as a more effective starting point for multimodal pre-training initialization.

% \textbf{Visual Attention Mask}. We discovered that using a causal mask on visual tokens yields better results than using a full mask. One possible explanation is that the causal mask better fits the inherent structure of LLM.
\textbf{Visual Attention Mask}. We empirically find that using a causal mask on visual tokens yields a better result in comparison with a full mask. This is slightly counterintuitive, as using a bidirectional attention mask allows access to more information than a causal mask. We hypothesize the possible explanation for this phenomenon is that the causal mask better fits the inherent structure of LLMs.

% \textbf{Image Loss}. We tried introducing visual self-supervised loss to pretraining. Specifically, we let each visual token predict the clip feature of the next position for visual self-supervision. Similar to PaLI-X's finding~\citep{chen2023pali}, introducing a visual self-supervised task did not enhance the model's performance on multi-modal tasks.
\textbf{Image SSL Loss}. We also investigated the self-supervised learning loss on image features, where each visual feature predicts the CLIP feature of the next position for visual self-supervision. Align with the observation from PaLI-X~\citep{chen2023pali}, we find it brings no improvement on downstream tasks, although we indeed observed improvements in small models in our early experiments.

\textbf{Visual Encoder}. we substituted the 300M-parameter EVA2-L model for the 4.4B-parameter EVA2-E to investigate the impact of visual encoder parameters on various tasks. The results indicated that there was only a slight decrease in performance across most benchmarks. However, a notable exception was observed in the text-oriented dataset TextVQA, where we recorded a decline of 2.5.

\textbf{EMA}.
We utilize EMA (Exponential Moving Average) during pretraining. The ablation results show that EMA often brings improvements across various tasks compared to not using it.
% In the realm of optimization, we gravitated towards the Adam-EMA as our optimizer, underpinning the model's agility and computational efficacy. Central to this approach was the gradient \( g_t \) at each iteration, instrumental in the recalibration of the parameter \( p \), articulated as:
% \begin{equation}
%     p_t = p_{t-1} - \alpha \times \text{Function}(g_t, m, v)
% \end{equation}
% In this context, the \(\text{Function}\) embodies a nuanced computation, delineating the synergy between the gradient, its first moment (mean), and its second moment (uncentered variance).

% In a further refinement, the assimilation of the Exponential Moving Average (EMA) furnished a consistent parameter-refinement technique, indispensable for model equanimity. The equation
% \begin{equation}
%     p_{\mathrm{ema\phantom{,}} t} = \gamma \times p_{\mathrm{ema\phantom{,}} t-1} + (1 - \gamma) \times p_t
% \end{equation}
% succinctly encapsulates the quintessence of EMA. Within this schema, \( p_{\text{ema}, t} \) denotes the EMA of the parameter at the \( t \) juncture. The decay coefficient, \( \gamma \), nestled between 0 and 1, adeptly adjusts the historical significance, with proximities to 1 amplifying the refinement.

% In conclusion, subsequent to our rigorous training cycle, we accorded precedence to the EMA-calibrated parameters, \( p_{\text{ema}} \), over the conventional Adam-modified parameters during both validation and evaluation stages. This deliberate choice stemmed from the discernible robustness and superior adaptability inherent to the former.

\section{Conclusion} 
In this paper, we introduce CogVLM, an open visual language foundation model. CogVLM shifts the paradigm for VLM training from shallow alignment to deep fusion, achieving state-of-the-art performance on 17 classic multi-modal benchmarks. 

The VLM training is still in its infancy, and there are many directions to explore, for example, better SFT alignment, RLHF and anti-hallucination. Since the previous famous VLMs are mostly closed-source, we believe CogVLM will be a solid foundation for future multi-modal research.

% \section{Impact Statements}
% This paper presents work whose goal is to advance the field of Multimodal Learning. There are many potential societal consequences of our work, none which we feel must be specifically highlighted here.

\nocite{langley00}

\bibliography{example_paper}

\begin{thebibliography}{73}
\providecommand{\natexlab}[1]{#1}
\providecommand{\url}[1]{\texttt{#1}}
\expandafter\ifx\csname urlstyle\endcsname\relax
  \providecommand{\doi}[1]{doi: #1}\else
  \providecommand{\doi}{doi: \begingroup \urlstyle{rm}\Url}\fi

\bibitem[Agrawal et~al.(2019)Agrawal, Desai, Wang, Chen, Jain, Johnson, Batra, Parikh, Lee, and Anderson]{agrawal2019nocaps}
Agrawal, H., Desai, K., Wang, Y., Chen, X., Jain, R., Johnson, M., Batra, D., Parikh, D., Lee, S., and Anderson, P.
\newblock Nocaps: Novel object captioning at scale.
\newblock In \emph{Proceedings of the IEEE/CVF international conference on computer vision}, pp.\  8948--8957, 2019.

\bibitem[Alayrac et~al.(2022)Alayrac, Donahue, Luc, Miech, Barr, Hasson, Lenc, Mensch, Millican, Reynolds, et~al.]{alayrac2022flamingo}
Alayrac, J.-B., Donahue, J., Luc, P., Miech, A., Barr, I., Hasson, Y., Lenc, K., Mensch, A., Millican, K., Reynolds, M., et~al.
\newblock Flamingo: a visual language model for few-shot learning.
\newblock \emph{Advances in Neural Information Processing Systems}, 35:\penalty0 23716--23736, 2022.

\bibitem[Antol et~al.(2015)Antol, Agrawal, Lu, Mitchell, Batra, Zitnick, and Parikh]{antol2015vqa}
Antol, S., Agrawal, A., Lu, J., Mitchell, M., Batra, D., Zitnick, C.~L., and Parikh, D.
\newblock Vqa: Visual question answering.
\newblock In \emph{Proceedings of the IEEE international conference on computer vision}, pp.\  2425--2433, 2015.

\bibitem[Awadalla et~al.(2023)Awadalla, Gao, Gardner, Hessel, Hanafy, Zhu, Marathe, Bitton, Gadre, Sagawa, Jitsev, Kornblith, Koh, Ilharco, Wortsman, and Schmidt]{awadalla2023openflamingo}
Awadalla, A., Gao, I., Gardner, J., Hessel, J., Hanafy, Y., Zhu, W., Marathe, K., Bitton, Y., Gadre, S., Sagawa, S., Jitsev, J., Kornblith, S., Koh, P.~W., Ilharco, G., Wortsman, M., and Schmidt, L.
\newblock Openflamingo: An open-source framework for training large autoregressive vision-language models.
\newblock \emph{arXiv preprint arXiv:2308.01390}, 2023.

\bibitem[Bai et~al.(2023)Bai, Bai, Yang, Wang, Tan, Wang, Lin, Zhou, and Zhou]{bai2023qwen}
Bai, J., Bai, S., Yang, S., Wang, S., Tan, S., Wang, P., Lin, J., Zhou, C., and Zhou, J.
\newblock Qwen-vl: A frontier large vision-language model with versatile abilities.
\newblock \emph{arXiv preprint arXiv:2308.12966}, 2023.

\bibitem[Bavishi et~al.(2023)Bavishi, Elsen, Hawthorne, Nye, Odena, Somani, and Ta\c{s}\i{}rlar]{fuyu-8b}
Bavishi, R., Elsen, E., Hawthorne, C., Nye, M., Odena, A., Somani, A., and Ta\c{s}\i{}rlar, S.
\newblock Introducing our multimodal models, 2023.
\newblock URL \url{https://www.adept.ai/blog/fuyu-8b}.

\bibitem[Byeon et~al.(2022)Byeon, Park, Kim, Lee, Baek, and Kim]{kakaobrain2022coyo700m}
Byeon, M., Park, B., Kim, H., Lee, S., Baek, W., and Kim, S.
\newblock Coyo-700m: Image-text pair dataset.
\newblock \url{https://github.com/kakaobrain/coyo-dataset}, 2022.

\bibitem[Cadene et~al.(2019)Cadene, Ben-Younes, Cord, and Thome]{cadene2019murel}
Cadene, R., Ben-Younes, H., Cord, M., and Thome, N.
\newblock Murel: Multimodal relational reasoning for visual question answering.
\newblock In \emph{Proceedings of the IEEE/CVF conference on computer vision and pattern recognition}, pp.\  1989--1998, 2019.

\bibitem[Chen et~al.(2023{\natexlab{a}})Chen, Zhang, Zeng, Zhang, Zhu, and Zhao]{chen2023shikra}
Chen, K., Zhang, Z., Zeng, W., Zhang, R., Zhu, F., and Zhao, R.
\newblock Shikra: Unleashing multimodal llm's referential dialogue magic.
\newblock \emph{arXiv preprint arXiv:2306.15195}, 2023{\natexlab{a}}.

\bibitem[Chen et~al.(2022{\natexlab{a}})Chen, Li, Saxena, Hinton, and Fleet]{chen2022generalist}
Chen, T., Li, L., Saxena, S., Hinton, G., and Fleet, D.~J.
\newblock A generalist framework for panoptic segmentation of images and videos.
\newblock \emph{arXiv preprint arXiv:2210.06366}, 2022{\natexlab{a}}.

\bibitem[Chen et~al.(2022{\natexlab{b}})Chen, Wang, Changpinyo, Piergiovanni, Padlewski, Salz, Goodman, Grycner, Mustafa, Beyer, et~al.]{chen2022pali}
Chen, X., Wang, X., Changpinyo, S., Piergiovanni, A., Padlewski, P., Salz, D., Goodman, S., Grycner, A., Mustafa, B., Beyer, L., et~al.
\newblock Pali: A jointly-scaled multilingual language-image model.
\newblock \emph{arXiv preprint arXiv:2209.06794}, 2022{\natexlab{b}}.

\bibitem[Chen et~al.(2023{\natexlab{b}})Chen, Djolonga, Padlewski, Mustafa, Changpinyo, Wu, Ruiz, Goodman, Wang, Tay, et~al.]{chen2023pali}
Chen, X., Djolonga, J., Padlewski, P., Mustafa, B., Changpinyo, S., Wu, J., Ruiz, C.~R., Goodman, S., Wang, X., Tay, Y., et~al.
\newblock Pali-x: On scaling up a multilingual vision and language model.
\newblock \emph{arXiv preprint arXiv:2305.18565}, 2023{\natexlab{b}}.

\bibitem[Chiang et~al.(2023)Chiang, Li, Lin, Sheng, Wu, Zhang, Zheng, Zhuang, Zhuang, Gonzalez, et~al.]{chiang2023vicuna}
Chiang, W.-L., Li, Z., Lin, Z., Sheng, Y., Wu, Z., Zhang, H., Zheng, L., Zhuang, S., Zhuang, Y., Gonzalez, J.~E., et~al.
\newblock Vicuna: An open-source chatbot impressing gpt-4 with 90\%* chatgpt quality.
\newblock \emph{See https://vicuna. lmsys. org (accessed 14 April 2023)}, 2023.

\bibitem[Cornia et~al.(2021)Cornia, Baraldi, Fiameni, and Cucchiara]{cornia2021universal}
Cornia, M., Baraldi, L., Fiameni, G., and Cucchiara, R.
\newblock Universal captioner: Long-tail vision-and-language model training through content-style separation.
\newblock \emph{arXiv preprint arXiv:2111.12727}, 1\penalty0 (2):\penalty0 4, 2021.

\bibitem[Dai et~al.(2023)Dai, Li, Li, Tiong, Zhao, Wang, Li, Fung, and Hoi]{dai2023instructblip}
Dai, W., Li, J., Li, D., Tiong, A. M.~H., Zhao, J., Wang, W., Li, B., Fung, P., and Hoi, S.
\newblock Instructblip: Towards general-purpose vision-language models with instruction tuning, 2023.

\bibitem[Dong et~al.(2023)Dong, Han, Peng, Qi, Ge, Yang, Zhao, Sun, Zhou, Wei, et~al.]{dong2023dreamllm}
Dong, R., Han, C., Peng, Y., Qi, Z., Ge, Z., Yang, J., Zhao, L., Sun, J., Zhou, H., Wei, H., et~al.
\newblock Dreamllm: Synergistic multimodal comprehension and creation.
\newblock \emph{arXiv preprint arXiv:2309.11499}, 2023.

\bibitem[Driess et~al.(2023)Driess, Xia, Sajjadi, Lynch, Chowdhery, Ichter, Wahid, Tompson, Vuong, Yu, et~al.]{driess2023palm}
Driess, D., Xia, F., Sajjadi, M.~S., Lynch, C., Chowdhery, A., Ichter, B., Wahid, A., Tompson, J., Vuong, Q., Yu, T., et~al.
\newblock Palm-e: An embodied multimodal language model.
\newblock \emph{arXiv preprint arXiv:2303.03378}, 2023.

\bibitem[Hendrycks et~al.(2020)Hendrycks, Burns, Basart, Zou, Mazeika, Song, and Steinhardt]{hendrycks2020measuring}
Hendrycks, D., Burns, C., Basart, S., Zou, A., Mazeika, M., Song, D., and Steinhardt, J.
\newblock Measuring massive multitask language understanding.
\newblock \emph{arXiv preprint arXiv:2009.03300}, 2020.

\bibitem[Honnibal \& Johnson(2015)Honnibal and Johnson]{honnibal2015improved}
Honnibal, M. and Johnson, M.
\newblock An improved non-monotonic transition system for dependency parsing.
\newblock In \emph{Proceedings of the 2015 conference on empirical methods in natural language processing}, pp.\  1373--1378, 2015.

\bibitem[Hu et~al.(2021)Hu, Shen, Wallis, Allen-Zhu, Li, Wang, Wang, and Chen]{hu2021lora}
Hu, E.~J., Shen, Y., Wallis, P., Allen-Zhu, Z., Li, Y., Wang, S., Wang, L., and Chen, W.
\newblock Lora: Low-rank adaptation of large language models.
\newblock \emph{arXiv preprint arXiv:2106.09685}, 2021.

\bibitem[Hu et~al.(2022)Hu, Gan, Wang, Yang, Liu, Lu, and Wang]{hu2022scaling}
Hu, X., Gan, Z., Wang, J., Yang, Z., Liu, Z., Lu, Y., and Wang, L.
\newblock Scaling up vision-language pre-training for image captioning.
\newblock In \emph{Proceedings of the IEEE/CVF conference on computer vision and pattern recognition}, pp.\  17980--17989, 2022.

\bibitem[Kafle \& Kanan(2017)Kafle and Kanan]{kafle2017analysis}
Kafle, K. and Kanan, C.
\newblock An analysis of visual question answering algorithms.
\newblock In \emph{Proceedings of the IEEE international conference on computer vision}, pp.\  1965--1973, 2017.

\bibitem[Kazemzadeh et~al.(2014)Kazemzadeh, Ordonez, Matten, and Berg]{kazemzadeh2014referitgame}
Kazemzadeh, S., Ordonez, V., Matten, M., and Berg, T.
\newblock Referitgame: Referring to objects in photographs of natural scenes.
\newblock In \emph{Proceedings of the 2014 conference on empirical methods in natural language processing (EMNLP)}, pp.\  787--798, 2014.

\bibitem[Krasin et~al.(2017)Krasin, Duerig, Alldrin, Ferrari, Abu-El-Haija, Kuznetsova, Rom, Uijlings, Popov, Veit, et~al.]{krasin2017openimages}
Krasin, I., Duerig, T., Alldrin, N., Ferrari, V., Abu-El-Haija, S., Kuznetsova, A., Rom, H., Uijlings, J., Popov, S., Veit, A., et~al.
\newblock Openimages: A public dataset for large-scale multi-label and multi-class image classification.
\newblock \emph{Dataset available from https://github. com/openimages}, 2\penalty0 (3):\penalty0 18, 2017.

\bibitem[Krishna et~al.(2017)Krishna, Zhu, Groth, Johnson, Hata, Kravitz, Chen, Kalantidis, Li, Shamma, et~al.]{krishna2017visual}
Krishna, R., Zhu, Y., Groth, O., Johnson, J., Hata, K., Kravitz, J., Chen, S., Kalantidis, Y., Li, L.-J., Shamma, D.~A., et~al.
\newblock Visual genome: Connecting language and vision using crowdsourced dense image annotations.
\newblock \emph{International journal of computer vision}, 123:\penalty0 32--73, 2017.

\bibitem[Laurençon et~al.(2023)Laurençon, Saulnier, Tronchon, Bekman, Singh, Lozhkov, Wang, Karamcheti, Rush, Kiela, Cord, and Sanh]{laurencon2023obelics}
Laurençon, H., Saulnier, L., Tronchon, L., Bekman, S., Singh, A., Lozhkov, A., Wang, T., Karamcheti, S., Rush, A.~M., Kiela, D., Cord, M., and Sanh, V.
\newblock Obelics: An open web-scale filtered dataset of interleaved image-text documents, 2023.

\bibitem[Li et~al.(2023{\natexlab{a}})Li, Wang, Wang, Ge, Ge, and Shan]{li2023seed}
Li, B., Wang, R., Wang, G., Ge, Y., Ge, Y., and Shan, Y.
\newblock Seed-bench: Benchmarking multimodal llms with generative comprehension.
\newblock \emph{arXiv preprint arXiv:2307.16125}, 2023{\natexlab{a}}.

\bibitem[Li et~al.(2023{\natexlab{b}})Li, Li, Savarese, and Hoi]{li2023blip}
Li, J., Li, D., Savarese, S., and Hoi, S.
\newblock Blip-2: Bootstrapping language-image pre-training with frozen image encoders and large language models.
\newblock \emph{arXiv preprint arXiv:2301.12597}, 2023{\natexlab{b}}.

\bibitem[Li et~al.(2023{\natexlab{c}})Li, Du, Zhou, Wang, Zhao, and Wen]{li2023evaluating}
Li, Y., Du, Y., Zhou, K., Wang, J., Zhao, W.~X., and Wen, J.-R.
\newblock Evaluating object hallucination in large vision-language models.
\newblock \emph{arXiv preprint arXiv:2305.10355}, 2023{\natexlab{c}}.

\bibitem[Lin et~al.(2023{\natexlab{a}})Lin, Yuan, Wu, Wang, and Wang]{lin2023uninext}
Lin, F., Yuan, J., Wu, S., Wang, F., and Wang, Z.
\newblock Uninext: Exploring a unified architecture for vision recognition.
\newblock \emph{arXiv preprint arXiv:2304.13700}, 2023{\natexlab{a}}.

\bibitem[Lin et~al.(2014)Lin, Maire, Belongie, Hays, Perona, Ramanan, Doll{\'a}r, and Zitnick]{lin2014microsoft}
Lin, T.-Y., Maire, M., Belongie, S., Hays, J., Perona, P., Ramanan, D., Doll{\'a}r, P., and Zitnick, C.~L.
\newblock Microsoft coco: Common objects in context.
\newblock In \emph{Computer Vision--ECCV 2014: 13th European Conference, Zurich, Switzerland, September 6-12, 2014, Proceedings, Part V 13}, pp.\  740--755. Springer, 2014.

\bibitem[Lin et~al.(2023{\natexlab{b}})Lin, Liu, Zhang, Gao, Qiu, Xiao, Qiu, Lin, Shao, Chen, et~al.]{lin2023sphinx}
Lin, Z., Liu, C., Zhang, R., Gao, P., Qiu, L., Xiao, H., Qiu, H., Lin, C., Shao, W., Chen, K., et~al.
\newblock Sphinx: The joint mixing of weights, tasks, and visual embeddings for multi-modal large language models.
\newblock \emph{arXiv preprint arXiv:2311.07575}, 2023{\natexlab{b}}.

\bibitem[Liu et~al.(2023{\natexlab{a}})Liu, Lin, Li, Wang, Yacoob, and Wang]{liu2023aligning}
Liu, F., Lin, K., Li, L., Wang, J., Yacoob, Y., and Wang, L.
\newblock Aligning large multi-modal model with robust instruction tuning.
\newblock \emph{arXiv preprint arXiv:2306.14565}, 2023{\natexlab{a}}.

\bibitem[Liu et~al.(2023{\natexlab{b}})Liu, Li, Li, and Lee]{liu2023improved}
Liu, H., Li, C., Li, Y., and Lee, Y.~J.
\newblock Improved baselines with visual instruction tuning.
\newblock \emph{arXiv preprint arXiv:2310.03744}, 2023{\natexlab{b}}.

\bibitem[Liu et~al.(2023{\natexlab{c}})Liu, Li, Wu, and Lee]{liu2023visual}
Liu, H., Li, C., Wu, Q., and Lee, Y.~J.
\newblock Visual instruction tuning.
\newblock \emph{arXiv preprint arXiv:2304.08485}, 2023{\natexlab{c}}.

\bibitem[Liu et~al.(2017)Liu, Wang, and Yang]{liu2017referring}
Liu, J., Wang, L., and Yang, M.-H.
\newblock Referring expression generation and comprehension via attributes.
\newblock In \emph{Proceedings of the IEEE International Conference on Computer Vision}, pp.\  4856--4864, 2017.

\bibitem[Liu et~al.(2023{\natexlab{d}})Liu, Fan, Johns, Yu, Xiao, and Anandkumar]{liu2023prismer}
Liu, S., Fan, L., Johns, E., Yu, Z., Xiao, C., and Anandkumar, A.
\newblock Prismer: A vision-language model with an ensemble of experts.
\newblock \emph{arXiv preprint arXiv:2303.02506}, 2023{\natexlab{d}}.

\bibitem[Liu et~al.(2023{\natexlab{e}})Liu, Zeng, Ren, Li, Zhang, Yang, Li, Yang, Su, Zhu, et~al.]{liu2023grounding}
Liu, S., Zeng, Z., Ren, T., Li, F., Zhang, H., Yang, J., Li, C., Yang, J., Su, H., Zhu, J., et~al.
\newblock Grounding dino: Marrying dino with grounded pre-training for open-set object detection.
\newblock \emph{arXiv preprint arXiv:2303.05499}, 2023{\natexlab{e}}.

\bibitem[Liu et~al.(2023{\natexlab{f}})Liu, Zheng, Du, Ding, Qian, Yang, and Tang]{liu2023gpt}
Liu, X., Zheng, Y., Du, Z., Ding, M., Qian, Y., Yang, Z., and Tang, J.
\newblock Gpt understands, too.
\newblock \emph{AI Open}, 2023{\natexlab{f}}.

\bibitem[Liu et~al.(2023{\natexlab{g}})Liu, Duan, Zhang, Li, Zhang, Zhao, Yuan, Wang, He, Liu, et~al.]{liu2023mmbench}
Liu, Y., Duan, H., Zhang, Y., Li, B., Zhang, S., Zhao, W., Yuan, Y., Wang, J., He, C., Liu, Z., et~al.
\newblock Mmbench: Is your multi-modal model an all-around player?
\newblock \emph{arXiv preprint arXiv:2307.06281}, 2023{\natexlab{g}}.

\bibitem[Lu et~al.(2022)Lu, Mishra, Xia, Qiu, Chang, Zhu, Tafjord, Clark, and Kalyan]{lu2022learn}
Lu, P., Mishra, S., Xia, T., Qiu, L., Chang, K.-W., Zhu, S.-C., Tafjord, O., Clark, P., and Kalyan, A.
\newblock Learn to explain: Multimodal reasoning via thought chains for science question answering.
\newblock \emph{Advances in Neural Information Processing Systems}, 35:\penalty0 2507--2521, 2022.

\bibitem[Lu et~al.(2023)Lu, Bansal, Xia, Liu, Li, Hajishirzi, Cheng, Chang, Galley, and Gao]{lu2023mathvista}
Lu, P., Bansal, H., Xia, T., Liu, J., Li, C., Hajishirzi, H., Cheng, H., Chang, K.-W., Galley, M., and Gao, J.
\newblock Mathvista: Evaluating mathematical reasoning of foundation models in visual contexts.
\newblock \emph{arXiv preprint arXiv:2310.02255}, 2023.

\bibitem[Mao et~al.(2016)Mao, Huang, Toshev, Camburu, Yuille, and Murphy]{mao2016generation}
Mao, J., Huang, J., Toshev, A., Camburu, O., Yuille, A.~L., and Murphy, K.
\newblock Generation and comprehension of unambiguous object descriptions.
\newblock In \emph{Proceedings of the IEEE conference on computer vision and pattern recognition}, pp.\  11--20, 2016.

\bibitem[Marino et~al.(2019)Marino, Rastegari, Farhadi, and Mottaghi]{marino2019ok}
Marino, K., Rastegari, M., Farhadi, A., and Mottaghi, R.
\newblock Ok-vqa: A visual question answering benchmark requiring external knowledge.
\newblock In \emph{Proceedings of the IEEE/cvf conference on computer vision and pattern recognition}, pp.\  3195--3204, 2019.

\bibitem[Mishra et~al.(2019)Mishra, Shekhar, Singh, and Chakraborty]{mishra2019ocr}
Mishra, A., Shekhar, S., Singh, A.~K., and Chakraborty, A.
\newblock Ocr-vqa: Visual question answering by reading text in images.
\newblock In \emph{2019 international conference on document analysis and recognition (ICDAR)}, pp.\  947--952. IEEE, 2019.

\bibitem[Peng et~al.()Peng, Wang, Dong, Hao, Huang, Ma, and Wei]{peng2023kosmos}
Peng, Z., Wang, W., Dong, L., Hao, Y., Huang, S., Ma, S., and Wei, F.
\newblock Kosmos-2: Grounding multimodal large language models to the world.
\newblock \emph{arXiv preprint arXiv:2306.14824}.

\bibitem[Plummer et~al.(2015)Plummer, Wang, Cervantes, Caicedo, Hockenmaier, and Lazebnik]{plummer2015flickr30k}
Plummer, B.~A., Wang, L., Cervantes, C.~M., Caicedo, J.~C., Hockenmaier, J., and Lazebnik, S.
\newblock Flickr30k entities: Collecting region-to-phrase correspondences for richer image-to-sentence models.
\newblock In \emph{Proceedings of the IEEE international conference on computer vision}, pp.\  2641--2649, 2015.

\bibitem[Raffel et~al.(2020)Raffel, Shazeer, Roberts, Lee, Narang, Matena, Zhou, Li, and Liu]{raffel2020exploring}
Raffel, C., Shazeer, N., Roberts, A., Lee, K., Narang, S., Matena, M., Zhou, Y., Li, W., and Liu, P.~J.
\newblock Exploring the limits of transfer learning with a unified text-to-text transformer.
\newblock \emph{The Journal of Machine Learning Research}, 21\penalty0 (1):\penalty0 5485--5551, 2020.

\bibitem[Schuhmann et~al.(2022)Schuhmann, Beaumont, Vencu, Gordon, Wightman, Cherti, Coombes, Katta, Mullis, Wortsman, et~al.]{schuhmann2022laion}
Schuhmann, C., Beaumont, R., Vencu, R., Gordon, C., Wightman, R., Cherti, M., Coombes, T., Katta, A., Mullis, C., Wortsman, M., et~al.
\newblock Laion-5b: An open large-scale dataset for training next generation image-text models.
\newblock \emph{Advances in Neural Information Processing Systems}, 35:\penalty0 25278--25294, 2022.

\bibitem[Shazeer(2020)]{shazeer2020glu}
Shazeer, N.
\newblock Glu variants improve transformer.
\newblock \emph{arXiv preprint arXiv:2002.05202}, 2020.

\bibitem[Shrestha et~al.(2019)Shrestha, Kafle, and Kanan]{shrestha2019answer}
Shrestha, R., Kafle, K., and Kanan, C.
\newblock Answer them all! toward universal visual question answering models.
\newblock In \emph{Proceedings of the IEEE/CVF conference on computer vision and pattern recognition}, pp.\  10472--10481, 2019.

\bibitem[Sidorov et~al.(2020)Sidorov, Hu, Rohrbach, and Singh]{sidorov2020textcaps}
Sidorov, O., Hu, R., Rohrbach, M., and Singh, A.
\newblock Textcaps: a dataset for image captioning with reading comprehension.
\newblock In \emph{Computer Vision--ECCV 2020: 16th European Conference, Glasgow, UK, August 23--28, 2020, Proceedings, Part II 16}, pp.\  742--758. Springer, 2020.

\bibitem[Singh et~al.(2019)Singh, Natarajan, Shah, Jiang, Chen, Batra, Parikh, and Rohrbach]{singh2019towards}
Singh, A., Natarajan, V., Shah, M., Jiang, Y., Chen, X., Batra, D., Parikh, D., and Rohrbach, M.
\newblock Towards vqa models that can read.
\newblock In \emph{Proceedings of the IEEE/CVF conference on computer vision and pattern recognition}, pp.\  8317--8326, 2019.

\bibitem[Sun et~al.(2023{\natexlab{a}})Sun, Cui, Zhang, Zhang, Yu, Luo, Wang, Rao, Liu, Huang, et~al.]{sun2023generative}
Sun, Q., Cui, Y., Zhang, X., Zhang, F., Yu, Q., Luo, Z., Wang, Y., Rao, Y., Liu, J., Huang, T., et~al.
\newblock Generative multimodal models are in-context learners.
\newblock \emph{arXiv preprint arXiv:2312.13286}, 2023{\natexlab{a}}.

\bibitem[Sun et~al.(2023{\natexlab{b}})Sun, Fang, Wu, Wang, and Cao]{sun2023eva}
Sun, Q., Fang, Y., Wu, L., Wang, X., and Cao, Y.
\newblock Eva-clip: Improved training techniques for clip at scale.
\newblock \emph{arXiv preprint arXiv:2303.15389}, 2023{\natexlab{b}}.

\bibitem[Touvron et~al.(2023)Touvron, Martin, Stone, Albert, Almahairi, Babaei, Bashlykov, Batra, Bhargava, Bhosale, et~al.]{touvron2023llama}
Touvron, H., Martin, L., Stone, K., Albert, P., Almahairi, A., Babaei, Y., Bashlykov, N., Batra, S., Bhargava, P., Bhosale, S., et~al.
\newblock Llama 2: Open foundation and fine-tuned chat models.
\newblock \emph{arXiv preprint arXiv:2307.09288}, 2023.

\bibitem[Tsimpoukelli et~al.(2021)Tsimpoukelli, Menick, Cabi, Eslami, Vinyals, and Hill]{tsimpoukelli2021multimodal}
Tsimpoukelli, M., Menick, J.~L., Cabi, S., Eslami, S., Vinyals, O., and Hill, F.
\newblock Multimodal few-shot learning with frozen language models.
\newblock \emph{Advances in Neural Information Processing Systems}, 34:\penalty0 200--212, 2021.

\bibitem[Wang et~al.(2022{\natexlab{a}})Wang, Yang, Hu, Li, Lin, Gan, Liu, Liu, and Wang]{wang2022git}
Wang, J., Yang, Z., Hu, X., Li, L., Lin, K., Gan, Z., Liu, Z., Liu, C., and Wang, L.
\newblock Git: A generative image-to-text transformer for vision and language.
\newblock \emph{arXiv preprint arXiv:2205.14100}, 2022{\natexlab{a}}.

\bibitem[Wang et~al.(2022{\natexlab{b}})Wang, Yang, Men, Lin, Bai, Li, Ma, Zhou, Zhou, and Yang]{wang2022ofa}
Wang, P., Yang, A., Men, R., Lin, J., Bai, S., Li, Z., Ma, J., Zhou, C., Zhou, J., and Yang, H.
\newblock Ofa: Unifying architectures, tasks, and modalities through a simple sequence-to-sequence learning framework.
\newblock In \emph{International Conference on Machine Learning}, pp.\  23318--23340. PMLR, 2022{\natexlab{b}}.

\bibitem[Wang et~al.(2023{\natexlab{a}})Wang, Wang, Lin, Bai, Zhou, Zhou, Wang, and Zhou]{wang2023one}
Wang, P., Wang, S., Lin, J., Bai, S., Zhou, X., Zhou, J., Wang, X., and Zhou, C.
\newblock One-peace: Exploring one general representation model toward unlimited modalities.
\newblock \emph{arXiv preprint arXiv:2305.11172}, 2023{\natexlab{a}}.

\bibitem[Wang et~al.(2023{\natexlab{b}})Wang, Chen, Chen, Wu, Zhu, Zeng, Luo, Lu, Zhou, Qiao, et~al.]{wang2023visionllm}
Wang, W., Chen, Z., Chen, X., Wu, J., Zhu, X., Zeng, G., Luo, P., Lu, T., Zhou, J., Qiao, Y., et~al.
\newblock Visionllm: Large language model is also an open-ended decoder for vision-centric tasks.
\newblock \emph{arXiv preprint arXiv:2305.11175}, 2023{\natexlab{b}}.

\bibitem[Wang et~al.(2021)Wang, Yu, Yu, Dai, Tsvetkov, and Cao]{wang2021simvlm}
Wang, Z., Yu, J., Yu, A.~W., Dai, Z., Tsvetkov, Y., and Cao, Y.
\newblock Simvlm: Simple visual language model pretraining with weak supervision.
\newblock \emph{arXiv preprint arXiv:2108.10904}, 2021.

\bibitem[Ye et~al.(2023)Ye, Xu, Ye, Yan, Liu, Qian, Zhang, Huang, and Zhou]{ye2023mplug}
Ye, Q., Xu, H., Ye, J., Yan, M., Liu, H., Qian, Q., Zhang, J., Huang, F., and Zhou, J.
\newblock mplug-owl2: Revolutionizing multi-modal large language model with modality collaboration.
\newblock \emph{arXiv preprint arXiv:2311.04257}, 2023.

\bibitem[You et~al.(2023)You, Zhang, Gan, Du, Zhang, Wang, Cao, Chang, and Yang]{you2023ferret}
You, H., Zhang, H., Gan, Z., Du, X., Zhang, B., Wang, Z., Cao, L., Chang, S.-F., and Yang, Y.
\newblock Ferret: Refer and ground anything anywhere at any granularity.
\newblock \emph{arXiv preprint arXiv:2310.07704}, 2023.

\bibitem[Yu et~al.(2022)Yu, Wang, Vasudevan, Yeung, Seyedhosseini, and Wu]{yu2022coca}
Yu, J., Wang, Z., Vasudevan, V., Yeung, L., Seyedhosseini, M., and Wu, Y.
\newblock Coca: Contrastive captioners are image-text foundation models.
\newblock \emph{arXiv preprint arXiv:2205.01917}, 2022.

\bibitem[Yu et~al.(2016)Yu, Poirson, Yang, Berg, and Berg]{yu2016modeling}
Yu, L., Poirson, P., Yang, S., Berg, A.~C., and Berg, T.~L.
\newblock Modeling context in referring expressions.
\newblock In \emph{Computer Vision--ECCV 2016: 14th European Conference, Amsterdam, The Netherlands, October 11-14, 2016, Proceedings, Part II 14}, pp.\  69--85. Springer, 2016.

\bibitem[Yu et~al.(2023)Yu, Yang, Li, Wang, Lin, Liu, Wang, and Wang]{yu2023mm}
Yu, W., Yang, Z., Li, L., Wang, J., Lin, K., Liu, Z., Wang, X., and Wang, L.
\newblock Mm-vet: Evaluating large multimodal models for integrated capabilities.
\newblock \emph{arXiv preprint arXiv:2308.02490}, 2023.

\bibitem[Yue et~al.(2023)Yue, Ni, Zhang, Zheng, Liu, Zhang, Stevens, Jiang, Ren, Sun, et~al.]{yue2023mmmu}
Yue, X., Ni, Y., Zhang, K., Zheng, T., Liu, R., Zhang, G., Stevens, S., Jiang, D., Ren, W., Sun, Y., et~al.
\newblock Mmmu: A massive multi-discipline multimodal understanding and reasoning benchmark for expert agi.
\newblock \emph{arXiv preprint arXiv:2311.16502}, 2023.

\bibitem[Zhang et~al.(2022)Zhang, Zhang, Hu, Chen, Li, Dai, Wang, Yuan, Hwang, and Gao]{zhang2022glipv2}
Zhang, H., Zhang, P., Hu, X., Chen, Y.-C., Li, L., Dai, X., Wang, L., Yuan, L., Hwang, J.-N., and Gao, J.
\newblock Glipv2: Unifying localization and vision-language understanding.
\newblock \emph{Advances in Neural Information Processing Systems}, 35:\penalty0 36067--36080, 2022.

\bibitem[Zhang et~al.(2021)Zhang, Li, Hu, Yang, Zhang, Wang, Choi, and Gao]{zhang2021vinvl}
Zhang, P., Li, X., Hu, X., Yang, J., Zhang, L., Wang, L., Choi, Y., and Gao, J.
\newblock Vinvl: Revisiting visual representations in vision-language models.
\newblock In \emph{Proceedings of the IEEE/CVF conference on computer vision and pattern recognition}, pp.\  5579--5588, 2021.

\bibitem[Zhang et~al.(2023)Zhang, Zhang, Gu, Zhou, Lipka, Yang, and Sun]{zhang2023llavar}
Zhang, Y., Zhang, R., Gu, J., Zhou, Y., Lipka, N., Yang, D., and Sun, T.
\newblock Llavar: Enhanced visual instruction tuning for text-rich image understanding, 2023.

\bibitem[Zhu et~al.(2023)Zhu, Chen, Shen, Li, and Elhoseiny]{zhu2023minigpt}
Zhu, D., Chen, J., Shen, X., Li, X., and Elhoseiny, M.
\newblock Minigpt-4: Enhancing vision-language understanding with advanced large language models.
\newblock \emph{arXiv preprint arXiv:2304.10592}, 2023.

\bibitem[Zhu et~al.(2016)Zhu, Groth, Bernstein, and Fei-Fei]{zhu2016visual7w}
Zhu, Y., Groth, O., Bernstein, M., and Fei-Fei, L.
\newblock Visual7w: Grounded question answering in images.
\newblock In \emph{Proceedings of the IEEE conference on computer vision and pattern recognition}, pp.\  4995--5004, 2016.

\end{thebibliography}
\bibliographystyle{icml2024}

%%%%%%%%%%%%%%%%%%%%%%%%%%%%%%%%%%%%%%%%%%%%%%%%%%%%%%%%%%%%%%%%%%%%%%%%%%%%%%%
%%%%%%%%%%%%%%%%%%%%%%%%%%%%%%%%%%%%%%%%%%%%%%%%%%%%%%%%%%%%%%%%%%%%%%%%%%%%%%%
% APPENDIX
%%%%%%%%%%%%%%%%%%%%%%%%%%%%%%%%%%%%%%%%%%%%%%%%%%%%%%%%%%%%%%%%%%%%%%%%%%%%%%%
%%%%%%%%%%%%%%%%%%%%%%%%%%%%%%%%%%%%%%%%%%%%%%%%%%%%%%%%%%%%%%%%%%%%%%%%%%%%%%%
\newpage
\appendix
\onecolumn
\appendix
\section{Appendix}
\subsection{Details of Training Settings} \label{app:A1}
We report the details of parameter settings during pre-training and multitask training in Table~\ref{tab:hyperparam_pretrain} and Table~\ref{tbl:ft:captioning:hyperparams}.
% We report the relevant parameters of module pre-training and multitask training in Table~\ref{tab:hyperparam_pretrain} and Table~\ref{tbl:ft:captioning:hyperparams}.

\begin{small}
\begin{table}[htbp]
\centering
    \renewcommand{\arraystretch}{1.15}
    \caption{
     Hyperparameters for pre-training model. 
  }
    \setlength{\tabcolsep}{0.35mm}{
    \begin{tabular}{c|cc}
    \toprule
    \bf Hyperparameters & \bf Stage 1 & \bf Stage 2\\
      \midrule
      Total steps & $120,000$  & $60,000$ \\
      Warmup steps & $12,000$  & $1,200$ \\
      Batch size & $8,192$  & $1,024$ \\
      Learning rate & $1e^{-4}$ & $1e^{-5}$ \\
      Learning rate decay & \multicolumn{2}{c}{$\operatorname{Cosine}$} \\
      Weight decay & \multicolumn{2}{c}{0.05} \\
      Dropout ratio & \multicolumn{2}{c}{$0.1$} \\
      Adam $\epsilon$ & \multicolumn{2}{c}{$1e^{-8}$}  \\
      Adam $\beta$ & \multicolumn{2}{c}{(0.9, 0.95)} \\
      \hline
      Textual encoder &  \multicolumn{2}{c}{Vicuna-1.5-7B}\\
      Visual encoder &  \multicolumn{2}{c}{EVA2-CLIP-E}\\
      Patch size & \multicolumn{2}{c}{$14$}  \\
      Input resolution & $224^2$ & $224^2 \rightarrow 490^2$  \\
      % \hline
    % Number of layers & $6$ \\
    %   Hidden size & $768$  \\
    %   FFN inner hidden size & $3072$  \\
    %   Number of attention heads & $12$  \\
    \bottomrule
    \end{tabular}}
    \vspace{2mm}
  \label{tab:hyperparam_pretrain}
\end{table}
\end{small}

\begin{table}[h]
\centering
\renewcommand{\arraystretch}{1.15}
\caption{
Hyperparameters for multitask finetuning \smodel.
}
\setlength{\tabcolsep}{1.2mm}{
\begin{tabular}{l|c}
\toprule
\bf Hyperparameters & \bf Multitask \\
\midrule
Learning rate & $1e^{-5}$ \\
Total steps & 6,000 \\
Batch size & 1,024 \\
AdamW $\epsilon$ & $1e^{-8}$  \\
AdamW $\beta$ & (0.9, 0.95) \\
Weight decay & 0.1 \\
Dropout ratio& 0.1 \\
Input resolution & $490^2$ \\
\bottomrule
\end{tabular}}
\vspace{2mm}
\label{tbl:ft:captioning:hyperparams}
\end{table}

\subsection{Details of Associated Datasets}
In this section, we introduce the details of datasets and their use in our evaluation process for all associated benchmarks.

\label{subsec:details_of_eval_benchmarks}
% Summary of Evaluation benchmarks

  \begin{table}[]
  \centering
\caption{Summary of the evaluation benchmarks.}
  \resizebox{0.98\textwidth}{!}{%
    \begin{tabular}{l|llll}
\toprule
\textbf{Task}                      & \textbf{Dataset} & \textbf{Description}                                & \textbf{Split}    & \textbf{Metrics}       \\ \midrule
\multirow{4}{*}{Image Caption}     & NoCaps           & Captioning of natural images.                       & val               & CIDEr ($\uparrow$)     \\
                                   & Flickr           & Captioning of natural images.                       & karpathy-test     & CIDEr ($\uparrow$)     \\
                                   & COCO             & Captioning of natural images.                       & karpathy-test     & CIDEr ($\uparrow$)     \\
                                   & TextCaps         & Captioning of natural images containing text.       & test              & CIDEr ($\uparrow$)     \\ \midrule
\multirow{4}{*}{General VQA}       & VQAv2            & VQA on natural images.                              & test-dev          & VQA Score($\uparrow$)  \\
                                   & OK-VQA           & VQA on natural images requiring outside knowledge.  & val               & VQA Score ($\uparrow$) \\
                                   & ScienceQA        & Multi-choice VQA on a diverse set of science topics & test              & Accuracy ($\uparrow$)  \\
                                   & TDIUC            & VQA on natural images with detailed question types. & val               & VQA Score ($\uparrow$) \\ \midrule
\multirow{2}{*}{Text-oriented VQA} & OCR-VQA          & VQA on images of book covers.                       & test              & EM ($\uparrow$)        \\
                                   & TextVQA          & VQA on natural images containing text.              & val               & VQA Score ($\uparrow$) \\ \midrule
\multirow{7}{*}{LVLM Benchmarks} & MM-Vet & Open-ended VQA on a diverse set of topics& test & GPT4 Score($\uparrow$) \\
&SEED-Bench & Multi-choice VQA on a diverse set of topics& IMG & Accuracy ($\uparrow$) \\
&MMBench & Multi-choice VQA on a diverse set of topics& test & Accuracy ($\uparrow$) \\
&LLaVA-Bench & Open-ended VQA for testing instruction following abilities&  In-the-Wild & GPT4 Score($\uparrow$)\\
&POPE & Multi-choice VQA for testing hallucinations & overall & Accuracy ($\uparrow$)\\
& MMMU & VQA on a diverse set of topics & test & Accuracy ($\uparrow$)\\
& MathVista & VQA for Measuring Mathematical Abilities & test-mini & Accuracy ($\uparrow$)\\
\midrule
\multirow{4}{*}{Grounding}         & RefCOCO          & Refer grounding on natural images.                  & overall            & Accuracy ($\uparrow$)  \\
                                   & RefCOCO+         & Refer grounding on natural images.                  & overall            & Accuracy ($\uparrow$)  \\
                                   & RefCOCOg         & Refer grounding on natural images.                  & overall              & Accuracy ($\uparrow$)  \\
                                   & Visual7W         & VQA with referential regions selection.             & val               & Accuracy ($\uparrow$) \\ \midrule
\end{tabular}
}
\label{tab:eval_benchmarks}
\end{table}

\subsubsection{Image Captioning}
\begin{itemize}
    \item \textbf{COCO}~\citep{lin2014microsoft} The Captions in COCO dataset are collected using Amazon’s Mechanical Turk (AMT) workers who are given instructions to control the quality. The dataset contains 330K images, where the train, validation and test sets contain 413,915 captions for 82,783 images, 202,520 captions for 40,504 images, and 379,249 captions for 40,775 images respectively.
    \item \textbf{NoCaps}~\citep{agrawal2019nocaps}. NoCaps is a large-scale benchmark for novel object captioning, containing nearly 400 novel object classes compared to COCO. The validation and test set comprised of 4,500 and 10,600 images, respectively, sourced from the Open Images~\citep{krasin2017openimages} and annotated with 11 human-generated captions per image, and each set is subdivided into three domains: ``in", ``near", and ``out", with objects in the ``out-domain" never appearing in the COCO dataset.
    \item \textbf{Flickr30K}~\citep{plummer2015flickr30k}. Flickr30K is a high-quality dataset consists of 31,783 images of everyday life activities, envets and scenes (all harvested from the online website Flickr) and 158,915 captions (obtained via crodsourcing). Each image in this dataset is described independently by five annotators who are not familiar with the specific entities and circumstances depicted in them.
    \item \textbf{TextCaps}~\citep{sidorov2020textcaps} Textcaps is a dataset with 145k captions for 28k images. The design purpose of the TextCaps dataset is to effectively integrate textual information with visual context into captions, requiring the model to have both excellent OCR capabilities and strong captioning abilities. 
\end{itemize}

\subsubsection{General VQA}
\begin{itemize}
    \item \textbf{VQAv2}~\citep{antol2015vqa} VQAv2 encompasses over 200,000 images, paired with more than 1.1 million questions that have collectively garnered over 11 million answers. Questions span various types, including yes/no, counting, and open-ended queries.
    \item \textbf{OKVQA}~\citep{marino2019ok} The OK-VQA (Outside Knowledge Visual Question Answering) dataset is specifically designed to probe visual question answering capabilities that necessitate external knowledge or common sense beyond image content. It has 14,055 open-ended questions and 5 ground truth answers per question.
    % \item \textbf{VizWiz-VQA}~\citep{gurari2018vizwiz} The VizWiz-VQA dataset is derived from blind individuals capturing images and voicing related questions, accompanied by 10 crowdsourced responses per query. The central challenge of this dataset involves predicting the visual question's answer and determining if it's unanswerable.
    \item \textbf{ScienceQA}~\citep{lu2022learn} The ScienceQA dataset comprises 21,208 multimodal multiple-choice questions spanning three diverse subjects: natural science, language science, and social science. Each question is annotated with explanations linked to relevant lectures.
    \item \textbf{TDIUC}~\citep{shrestha2019answer} The TDIUC dataset features 1.6M questions across 170K images from MS COCO and Visual Genome. Categorized into 12 distinct question types, it ranges from basic tasks like identifying objects or colors to more advanced reasoning like counting or positional discernment.
\end{itemize}

\subsubsection{Text-oriented VQA}
\begin{itemize}
    \item \textbf{OCRVQA}~\citep{mishra2019ocr} OCR-VQA consists of 207,572 book cover images with over 1 million question-answer pairs.
    \item \textbf{TextVQA}~\citep{singh2019towards} TextVQA is a dataset with 45,336 questions on 28,408 images that challenges models to detect, read, and reason about text within images to provide answers.
\end{itemize}

\subsection{LVLM Benchmarks}
\begin{itemize}
    \item \textbf{MM-Vet}~\cite{yu2023mm} MM-Vet defines six core VL capabilities and examines 16 integrations of interest derived from the combinations of these capabilities. It employs an evaluator based on LLMs for open-ended outputs, capable of assessing across different question types and answer styles, thus deriving a unified scoring metric.
    \item  \textbf{SEED-Bench}~\cite{li2023seed} SEED-Bench is a dataset comprising 19K multiple-choice questions with precise human annotations, covering 12 evaluation dimensions, including understanding of image and video modalities. It obtains accurate answer options through manual annotations, enabling objective and efficient assessment of model performance.
    \item \textbf{MMBench}~\cite{liu2023mmbench} MMBench comprises approximately 3000 multiple-choice questions, covering 20 different capability dimensions, aimed at evaluating various abilities of visual-language models. MMBench adopts a hierarchical capability dimension structure, including two high-level capability dimensions: perception and reasoning, as well as fine-grained capability dimensions such as object localization and attribute inference.
    \item \textbf{LLaVA-Bench}~\cite{liu2023visual} LLaVA-Bench (In-the-Wild) is a benchmark dataset comprising 60 questions, designed to evaluate the multimodal instruction following capabilities of LMMs. It includes indoor and outdoor scenes, memes, paintings, sketches, etc., and is equipped with highly detailed, manually curated descriptions and appropriate question selections.
    \item \textbf{POPE}~\cite{li2023evaluating} The POPE dataset is a binary classification query dataset specifically designed to evaluate object hallucination issues in LMMs. The random, popular, and adversarial subsets within the POPE dataset are constructed through different sampling strategies, totaling 8,910 entries.
    \item \textbf{MMMU}~\cite{yue2023mmmu} The MMMU dataset is a large-scale, multidisciplinary multimodal understanding and reasoning benchmark set, containing 11.5K questions. It covers 6 major disciplines, 30 topics, and 183 subfields, with question types including multiple-choice and open-ended questions. The dataset includes 30 types of images, such as charts, tables, chemical structures, photographs, paintings, musical scores, etc., testing the multimodal perception capabilities of models and their performance in expert-level tasks.
    \item  \textbf{MathVista}~\cite{lu2023mathvista} MathVista is a new benchmark dataset that combines mathematical and visual understanding, comprising 31 existing multimodal datasets and 3 newly created datasets, totaling 6141 examples. These datasets encompass a diverse range of mathematical reasoning abilities, including seven types: algebra, arithmetic, geometry, logic, numerical common sense, science, and statistics. The goal is to comprehensively evaluate the capabilities of existing foundational models in mathematical reasoning and visual understanding.

\end{itemize}

\subsubsection{Grounding}
\begin{itemize}
    \item \textbf{RefCOCO/RefCOCO+}~\citep{liu2017referring} RefCOCO and RefCOCO+ evolved from the ReferItGame. Both subsets focus on images with two or more similar objects. RefCOCO, with 142,209 expressions across 19,994 images, places no linguistic constraints. Conversely, RefCOCO+ emphasizes appearance-centric descriptions, omitting locational terms, and comprises 141,564 expressions over 19,992 images.
    \item \textbf{RefCOCOg}~\cite{mao2016generation} The RefCOCOg subset was amassed through Amazon Mechanical Turk, where workers penned natural referring expressions for objects in MSCOCO images; it boasts 85,474 referring expressions spanning 26,711 images, each containing 2 to 4 objects of the same category.
    \item \textbf{Visual7W}~\citep{zhu2016visual7w}. The Visual7W dataset is predominantly designed for VQA tasks, with a dedicated subset crafted for grounded VQA. In this subset, models are presented with an image accompanied by a ``which"-type question, such as ``Which is the small computer in the corner?". Participants are then given four bounding boxes within the image, from which they must select the correct one as the answer. The grounded Visual7W part consists of 25,733 images and 188,068 questions.
    \item \textbf{Flickr30K-Entities}~\citep{plummer2015flickr30k}. The Flickr30K Entities dataset, a precursor in the realm of grounded captioning, encompasses a collection of 31,783 images accompanied by 158k captioning annotations. Every caption in this dataset has been meticulously annotated such that each noun phrase is linked with a manually delineated referential bounding box. In total, there are 276k such annotated bounding boxes provided within this dataset.
    \item \textbf{VisualGenome~\citep{krishna2017visual}.} The VisualGenome dataset stands as a cornerstone in understanding the multifaceted relationships present within images. With a collection of over 100k images, each image is annotated in detail, capturing an average of 21 objects, 18 attributes, and 18 inter-object relationships. A unique aspect of this dataset is the alignment of objects, attributes, relationships, and region descriptions with standardized terminologies from WordNet. Specifically tailored for the REG and REC tasks, each annotated region in an image comes with a corresponding descriptive text, making it a rich resource for image understanding and semantic modeling. We use the subset with around 86k images and 3.6 million region-caption pairs for visual grounding.
\end{itemize}

% \subsection{Additional Fine-grained Experiments}

% \begin{wrapfigure}{l}{0.5\textwidth}
%   \begin{center}
%     \centering
%     \includegraphics[scale=0.5]{figures/tdiuc.pdf}
%     \caption{Performance on TDIUC benchmark with fine-grained questions classes.}
%     \label{fig:perform_on_tdiuc}
%   \end{center}
% \end{wrapfigure}

% To comprehensively investigate the proposed model on specific topics and question types, we further conduct extensive experiments on a representative benchmark, TDIUC~\citep{kafle2017analysis}.
% We use the publicly available split of val set as evaluation data, and the VQA accuracy calculated from their official scripts as evaluation metric.

% The experimental results on TDIUC, we compare our model against the specialist SOTA method Merel~\citep{cadene2019murel}, and the result is shown in Figure~\ref{fig:perform_on_tdiuc}. From the experimental result we can see that our model consistently outperforms the previous model on 12 specific question types, resulting a $94.0$ accuracy score compared to the existing SOTA of $88.2$ on the overall dataset. These results demonstrate that our model exhibits the comprehensive problem-solving skills on general VQA tasks.

\begin{figure*}[ht]
    \begin{center}
        \includegraphics[width=0.6\textwidth]{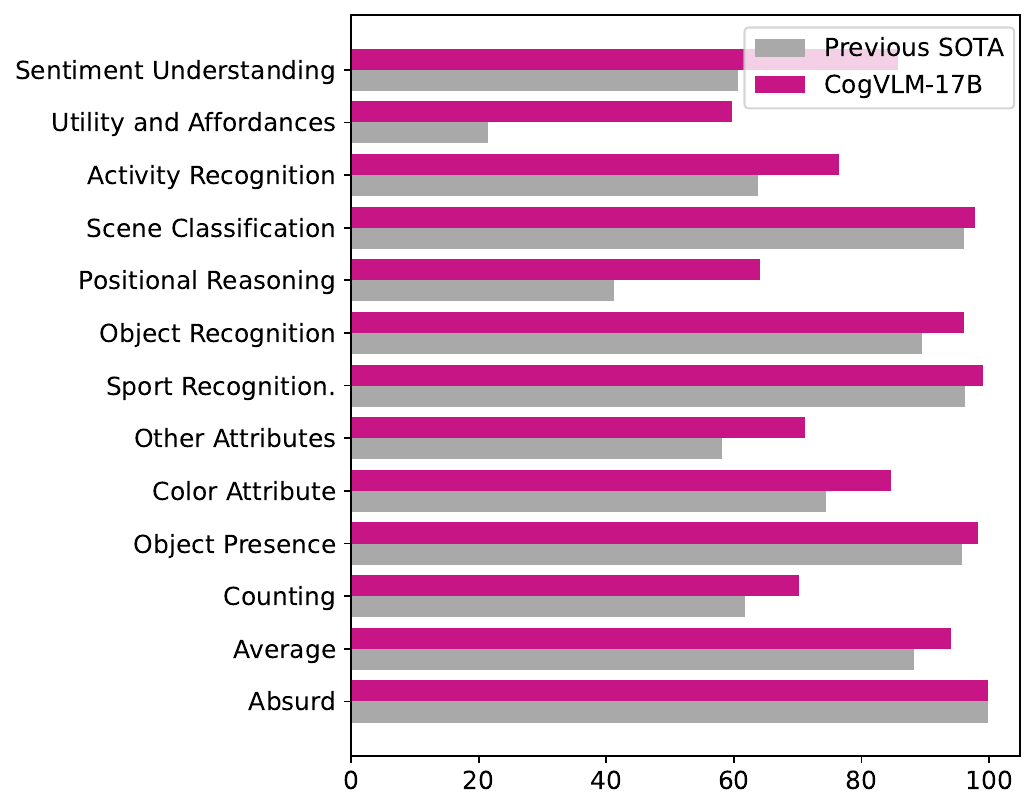} % Adjust the width as needed
        \caption{Performance on TDIUC benchmark with fine-grained questions classes.}
        \label{fig:perform_on_tdiuc}
    \end{center}
\end{figure*}

\section{Additional Fine-grained Experiments}

% This starts a figure environment that spans over the entire page width

To comprehensively investigate the proposed model on specific topics and question types, we further conduct extensive experiments on a representative benchmark, TDIUC~\citep{kafle2017analysis}.
We use the publicly available split of val set as evaluation data, and the VQA accuracy calculated from their official scripts as the evaluation metric.

The experimental results on TDIUC compare our model against the specialist SOTA method MUREL~\citep{cadene2019murel} are shown in Figure~\ref{fig:perform_on_tdiuc}. From the experimental result, we can see that our model consistently outperforms the previous model on 12 specific question types, resulting in a $94.0$ accuracy score compared to the previous SOTA of $88.2$ on the overall dataset. These results demonstrate that our model exhibits comprehensive problem-solving skills on general VQA tasks.

\section{Computational Efficiency}
In this section, we compare the computational efficiency of our model with other state-of-the-art models, considering both pretraining and finetuning data from datasets such as VQAv2 and TextVQA. Owing to an optimized architecture and the utilization of high-quality pretraining data, our model demonstrates a marked reduction in resource consumption during training relative to models with comparable parameter magnitudes.
\begin{table}[htbp]
\caption{Comparison of different models based on their computational efficiency. We use PFLOPS*days as metrics.}
\resizebox{\textwidth}{!}{%
\centering
\begin{tabular}{lcccc}
\toprule
\textbf{Model} & \textbf{Pretraining Data} & \textbf{Pretraining compute}& \textbf{VQAv2 finetuning}& \textbf{TextVQA finetuning} \\
\midrule
PaLI-3B & 1.6B & 56 & 1.1 & 0.2 \\
PaLI-17B & 1.6B & 453 & 4.5 & 0.9 \\
Flamingo-80B & 2.3B & 1381* & N/A & N/A \\
GIT2-5.1B & 12.9B & 5513* & N/A & N/A \\
\midrule
% (3.4*10^9 + 4.8*10^9)*360*180000*8192
CogVLM & 1.5B & 230.1 & 1.2 & 0.13\\
\bottomrule
\end{tabular}}
\end{table}
%%%%%%%%%%%%%%%%%%%%%%%%%%%%%%%%%%%%%%%%%%%%%%%%%%%%%%%%%%%%%%%%%%%%%%%%%%%%%%%
%%%%%%%%%%%%%%%%%%%%%%%%%%%%%%%%%%%%%%%%%%%%%%%%%%%%%%%%%%%%%%%%%%%%%%%%%%%%%%%

\end{document}